\documentclass[letters,journal]{IEEEtran}
\IEEEoverridecommandlockouts
\usepackage{cite}
\usepackage{amsmath,amssymb,amsfonts}
\usepackage{algorithmic}
\usepackage{graphicx}
\usepackage{textcomp}
\usepackage{xcolor}
\usepackage{algorithm}
\usepackage{amssymb}
\usepackage{multirow}
\usepackage{adjustbox}
\usepackage{placeins}
\usepackage{lipsum}
\usepackage{url}
\usepackage{soul,color}
\usepackage{makecell}

\begin{document}

\title{Overcoming Dynamic Environments: A Hybrid Approach to Motion Planning for Manipulators}

% 1. “Hybrid Motion Planning: An Integration of Sampling-Based Motion Planners and Velocity Potential Function for Robotic Manipulators”
% 2. “Overcoming Dynamic Environments: A Hybrid Approach to Motion Planning for Robotic Manipulators”
% 3. “Navigating the Unpredictable: A Study on Hybrid Motion Planning for Robotic Manipulators in Dynamic Environments”
% 4. “Bridging the Gap: A Hybrid Approach Leveraging SBMPs and VPF for Enhanced Motion Planning in Robotic Manipulators”
% 5. “Adapting to Change: The Role of Hybrid Motion Planning in Robotic Manipulators Operating in Dynamic Environments”
% 6. “Innovation in Motion Planning: The Hybridization of SBMPs and VPF for Robotic Manipulators”
% 7. “Exploring Hybrid Motion Planning: Enhancing Robotic Manipulator Performance in Dynamic Environments”
% 8. “The Future of Robotic Manipulation: A Study on Hybrid Motion Planning in Dynamic and Uncertain Environments”

% Gavin (30/6/2024): This is the correct author layout for IEEE T-ASE (although I am not sure about the emails since it is not in the template, but I assume this is ok.
\author{Ho Minh Quang Ngo,
        Dac Dang Khoa Nguyen,
        Dinh Tung Le,~\IEEEmembership{Student Member,~IEEE,}
        and Gavin Paul,~\IEEEmembership{Member,~IEEE}% <-this % stops a space
\thanks{All authors are with the UTS: Robotics Institute, University of Technology Sydney, Australia (e-mail: \{HoMinhQuang.Ngo, Khoa.Nguyen, DinhTung.Le, Gavin.Paul\}@uts.edu.au).}% <-this % stops a space
\thanks{Manuscript received July 1, 2024.}}

\maketitle

\begin{abstract}
Robotic manipulators operating in dynamic and uncertain environments require efficient motion planning to navigate obstacles while maintaining smooth trajectories. Velocity Potential Field (VPF) planners offer real-time adaptability but struggle with complex constraints and local minima, leading to suboptimal performance in cluttered spaces. Traditional approaches rely on pre-planned trajectories, but frequent recomputation is computationally expensive. 
This study proposes a hybrid motion planning approach, integrating an improved VPF with a Sampling-Based Motion Planner (SBMP). The SBMP ensures optimal path generation, while VPF provides real-time adaptability to dynamic obstacles. This combination enhances motion planning efficiency, stability, and computational feasibility, addressing key challenges in uncertain environments such as warehousing and surgical robotics.
\end{abstract}

\def\abstractname{Note to Practitioners}
\begin{abstract}
Ensuring safe and efficient motion planning for robotic manipulators in dynamic environments is a critical challenge in factories, warehouses, and surgical robotics. Reactive planners like VPF provide real-time obstacle avoidance, but they suffer from local minima and instability, potentially causing robots to become trapped. This research proposes a hybrid approach combining SBMP's optimal trajectory planning with VPF’s real-time adaptability, enabling robots to track pre-planned paths while efficiently reacting to obstacles. The approach is validated on real manipulators using a motion capture system, demonstrating improved performance and reduced computational overhead. Future research will explore applications in human-robot collaboration for safer and more efficient shared workspaces.
\end{abstract}


\section{Introduction}
Motion planning is essential for robotic arm operations, with the choice of a planner depending on environmental conditions. In static environments, sampling-based motion planners (SBMPs) such as Probabilistic Roadmaps (PRM) and Rapidly-Exploring Random Trees (RRT)\cite{kavraki-1996, kuffner-2001} are widely used for high-dimensional path planning. Their asymptotically optimal variants, PRM* and RRT*\cite{karaman-2011}, further enhance path quality by minimising cost functions\cite{Clifton2008}. Recent advancements have improved SBMP efficiency by reducing computational cost, memory usage, and path smoothness issues\cite{yuan-2020, Munasinghe2020, jiang-2022, yi-2022}.

An alternative approach is optimization-based motion planning, which generates smooth, collision-free trajectories while optimising predefined criteria. CHOMP\cite{zucker-2013}, TrajOpt\cite{schulman-2014}, and GPMP2\cite{mukadam-2018} provide solutions for structured environments where obstacles and constraints are well defined.

In dynamic environments, where obstacles move unpredictably, real-time adaptability is crucial. To address this, RT-RRT*\cite{naderi-2015} extends RRT* with a fixed time budget and continuously updated tree structure, enabling real-time replanning. Similarly, Online RRT* and Online FMT*\cite{chandler-2017} allow rapid trajectory adjustments in changing conditions. Further enhancements, such as integrating Kalman filtering and combining RRT with D*\cite{otte-2015, chen-2018, yuan-2023, hadzic-2023}, improve robustness in dynamic settings. Optimization-based methods have also evolved to support real-time trajectory optimization\cite{kramer-2020} and precise path-following of end-effectors\cite{kang-2020}.

\begin{figure}%[htbp]
\centerline{\includegraphics[width=0.95\linewidth]{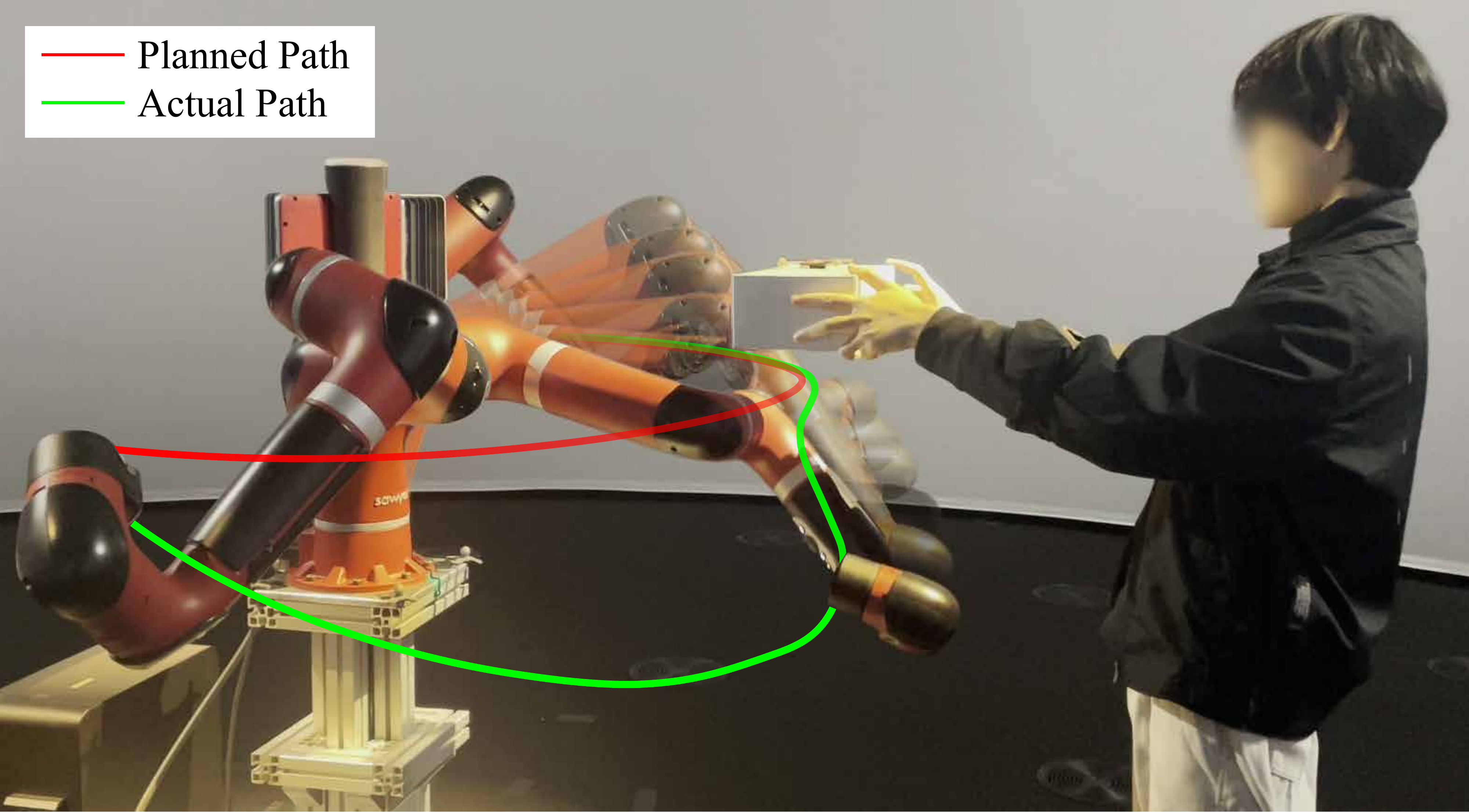}}
\caption{Trajectory executed on a Sawyer robot with a static obstacle}
\label{fig:ex_real_annotated_0}
\end{figure}

% Artificial Potential Functions (APF)\cite{khatib-2005, warren-2003} present a computationally lighter alternative. Recent research has addressed APF weaknesses such as local minima and singularities\cite{wang-2018, chen-2019, soodmand-2022}. Modern APF variants incorporate machine learning and advanced control techniques\cite{li-2021, zhu-2023, bai-2023} to handle dynamic environments effectively.

A different paradigm, Artificial Potential Functions (APF)\cite{khatib-2005, warren-2003}, offers computational efficiency but suffers from local minima and oscillatory behavior\cite{wang-2018, chen-2019, soodmand-2022}. Recent APF-based methods integrate machine learning\cite{li-2021} and predictive control\cite{zhu-2023, bai-2023} to improve adaptability. However, both SBMPs and APFs require extensive environmental knowledge, which is challenging in uncertain environments. A branch of APF, the Velocity Potential Function (VPF)\cite{cho-2002} presents a promising alternative by requiring less global information. Although improvements have been made\cite{xu-2018, xia-2023}, VPF alone struggles with suboptimal paths, goal convergence, and oscillatory motion in complex environments.

While real-time variants of SBMPs continuously update their search trees, they demand substantial computational resources. Similarly, replanning with optimization-based approaches in response to environmental changes incurs significant computational costs. A more efficient approach would combine the strengths of different planning paradigms while minimising their weaknesses.

This study proposes a hybrid approach integrating an SBMP with a VPF planner to address challenges in dynamic and unknown environments. The hybrid planner acquires global environmental information only once to establish an initial joint space path using SBMP, then employs VPF for real-time local adaptations and obstacle avoidance. This significantly reduces computational burden while maintaining navigation capabilities, particularly in environments where obstacles move unpredictably but the overall structure remains relatively stable. The study also introduces methods to seamlessly integrate VPF movements with the preplanned path, ensuring smooth transitions and adherence to motion constraints while addressing VPF's trapping issues and singularity risks. By combining SBMP's robustness in high-dimensional planning with VPF's simplicity and adaptability, the hybrid planner offers a comprehensive solution for manipulators operating in complex environments such as warehouse automation, surgical robotics, and other applications requiring navigation in dynamic, uncertain settings.

\section{Hybrid Planning for Dynamic Environments}
\subsection{Global Path in Joint Space}

The initial objective is to select an efficient global planner to navigate the manipulator through dynamic environments. Exploring paths directly within the manipulator's joint space was decided to simplify mapping complexities between the end effector's configuration space and the manipulator's joint space. In this study, this approach simplifies the task space - configuration space mapping complexity and enables smooth interpolation between manipulator joint configurations. This, in turn, facilitates the creation of smooth trajectories using well-known trajectory generators, like the quintic polynomial trajectory or trapezoidal velocity profile.

Given the dynamic nature of the goal environments, the speed of a global planner and its ability to explore free space are crucial. A slow planner may produce obsolete paths if it does not account for the current obstacles. In the worst case, it may never find a path in a changing environment. Therefore, while any SBMP can serve as a global planner, faster and more optimal SBMPs like RRT-Connect or RRT*~\cite{kuffner-2002, salzman-2016} are preferable in dynamic environments.

\begin{figure}%[htbp]
\centerline{\includegraphics[width=0.95\linewidth]{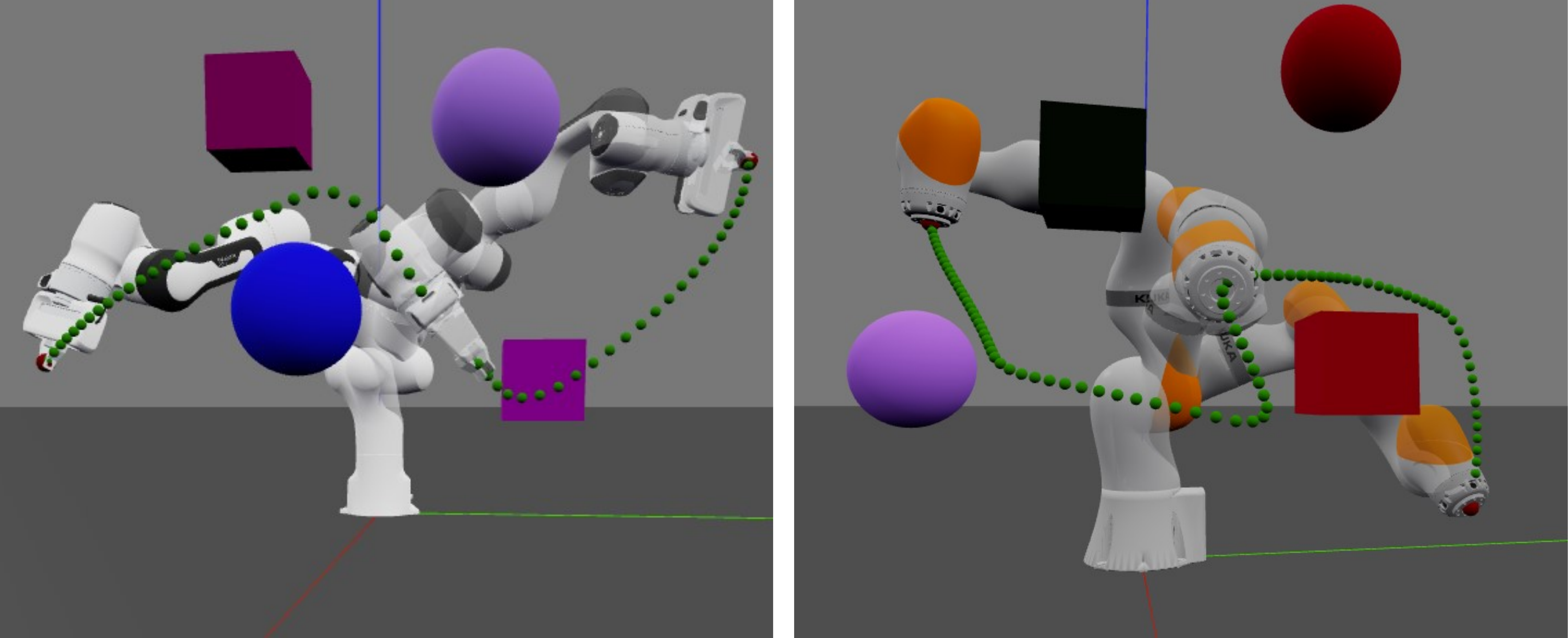}}
\caption{RRT* search in joint space for 7 DOF manipulators navigating through cluttered environments: the shortest paths to goal with the cost function is the Euclidean distance between joint configurations. Green dots indicate the positions of the end effector along the global paths with a fixed time step, representing the motion's smoothness.}
\label{p0}
\end{figure}

The inherent limitation of a sampling-based planner is its difficulty in managing continuous constraints, leading to a generally rough output. To mitigate this, a B-spline curve can be fitted to the waypoints from the raw path to output a global path with smooth transitions between path segments. This geometric path can be time parameterised to generate a trajectory subject to the motion constraints. With the pre-planned path in the joint space, the global path of $N-$ steps can be demonstrated by $\mathbf{Q} = \{\mathbf{q}_1, \mathbf{q}_2 ... \mathbf{q}_N\}$, where $\mathbf{q}$ is the joint configuration after a time step $\Delta t$, as an $n$-element vector for $n$ joint angles, i.e. $\mathbf{q} = [q_{1}, q_{2} ... q_{n}]$. Here, $\mathbf{q}_1$ and $\mathbf{q}_N$ represent the initial and goal joint configurations, respectively. Fig.~\ref{p0} demonstrates the time-parameterised paths for two manipulators from start to finish positions.  

%\FloatBarrier

\subsection{Global Path Tracking}
\label{sec:Global_Path_Tracking}

A purely reactive VPF-based approach, while simpler, may struggle with global optimality, goal convergence, and potential oscillations in complex environments. Therefore, the need of a global path-tracking controller is crucial for achieving a more complete, stable, and efficient robot control system. After the geometric path is time-parameterised into a trajectory $\mathbf{Q}$, a Proportional-Derivative (PD) controller for joint velocity control in its most basic form to follow that trajectory is:
\begin{equation}
    \label{js2}
    \mathbf{\dot{q}} = \mathbf{K}_P(\mathbf{q}_d(t) - \mathbf{q}(t)) + \mathbf{K}_D(\mathbf{\dot{q}}_d(t) - \mathbf{\dot{q}}(t))      
\end{equation}
where $\mathbf{K}_P$, $\mathbf{K}_D$ are the gain matrices,   $\mathbf{q}_d(t)$, $\mathbf{q}(t)$ and $\mathbf{\dot{q}}_d(t)$, $\mathbf{\dot{q}}(t)$ are the desired and actual joint position/velocity at time $t$, respectively, and $\mathbf{\dot{q}}$ is the joint velocity output.

For trajectory optimisation-based planners like TrajOpt or CHOMP, the trajectory should be re-planned in real time for environmental changes. Instead of replanning, this study proposes a local controller to drive the manipulator with a reaction velocity field to avoid moving obstacles and switches to a global controller to guide the manipulator to the global path when the field effect is negligible. The local controller will be discussed in the following sections, while the global controller is similar to a pure pursuit path tracking concept, where it tries to guide the manipulator back after the deviation from the global path caused by the reaction to obstacles' movements. The controller can be modified from \eqref{js2} as:
\begin{equation}
    \label{js3}
    \mathbf{\dot{q}} = \mathbf{K}_P(\mathbf{q}_{next} - \mathbf{q}(t)) + \mathbf{K}_D\mathbf{\dot{q}}_e
\end{equation}
where $\mathbf{q}_{next}$ is the look-ahead position in the path $\mathbf{Q}$ from the current position $\mathbf{q}(t)$, and $\mathbf{\dot{q}}_e$ is the derivative over time of the joint configuration error. 

For tracking the path, a look-ahead configuration $\mathbf{q}_{next} \in \mathbf{Q}$ can be obtained using a look-ahead step $s \in \mathbb{N}$. Given $\mathbf{q}_x \in \mathbf{Q}$ ($x \in \mathbb{N}$) is the closest configuration in the path, $\mathbf{q}_{next}$ can be chosen as $\mathbf{q}_y \in \mathbf{Q}$ that is $s$ steps away from $\mathbf{q}_x$ but not exceed the path length $N$:
\begin{equation}
    \label{js4}
    \mathbf{q}_{next} = \mathbf{q}_y \quad, y = \mathrm{min}(x + s, N)    
\end{equation}

The look-ahead step $s$ plays a crucial role in the path-tracking performance. Instead of a fixed step, a dynamic look-ahead selection can be applied based on two main factors: the current joint velocity's magnitude and the local path's curvature represented by a radian angle $\kappa \in [0, \pi]$. This approach allows the path-tracking controller to adapt to the path's local properties and the manipulator's current state. When it is moving quickly or the path is relatively straight, the planner looks further ahead, promoting smoother motion. Conversely, when the manipulator is moving slowly or the path has a high curvature, the planner looks closer ahead, ensuring more accurate tracking.  
%
% \begin{algorithm}
%     \caption{$\mathrm{approx\_curvature}$}
%     \label{al:js0}
%     \textbf{Input}:  $\mathbf{q}_1$, $\mathbf{q}_2$, $\mathbf{q}_3$ \\
%     \textbf{Output}: $\kappa$
%     \begin{algorithmic}[1]
%         \STATE $\kappa \leftarrow \arccos(\frac{(\mathbf{q}_3 - \mathbf{q}_2)\cdot(\mathbf{q}_2 - \mathbf{q}_1)}{\|\mathbf{q}_3 - \mathbf{q}_2\|\|\mathbf{q}_2 - \mathbf{q}_1\|})$
%         \RETURN $\kappa$
%     \end{algorithmic}
% \end{algorithm}
%
\begin{algorithm}
    % \caption{$\mathrm{get\_next\_config}$}
    \caption{$\text{Get the next  configuration in the path}$}
    \label{al:js1}
    \textbf{Input}:  $\mathbf{Q}$, $\mathbf{q}$, $\mathbf{\dot{q}}$, $k_v$, $k_c$ \\
    \textbf{Output}: $\mathbf{q}_{next}$
    \begin{algorithmic}[1]
    \STATE $\mathbf{q}_x, x \leftarrow \mathrm{get\_nearest\_config}
    (\mathbf{q}, \mathbf{Q})$
    \IF{$x < N-1$}
        % \STATE $\kappa \leftarrow \mathrm{approx\_curvature}(\mathbf{q}_{x-1}, \mathbf{q}_x, \mathbf{q}_{x+1})$
        \STATE $\kappa \leftarrow \arccos(\frac{(\mathbf{q}_{x+1} - \mathbf{q}_x)\cdot(\mathbf{q}_x - \mathbf{q}_{x-1})}{\|\mathbf{q}_{x+1} - \mathbf{q}_x\|\|\mathbf{q}_{x} - \mathbf{q}_{x-1}\|})$
        \STATE $s \leftarrow \mathrm{int}(k_v \|\mathbf{\dot{q}}\| + k_c\kappa + s_{base})$
        \STATE $s_{max} \leftarrow \min(s_{max}, N-x)$
        \STATE $s \leftarrow \min(s_{max},\max(s_{min}, s))$
    \ELSE
        \STATE $s \leftarrow 0$
    \ENDIF
    \STATE $\mathbf{q}_{next} \leftarrow \mathbf{q}_{x+s}$
    \RETURN $\mathbf{q}_{next}$
    \end{algorithmic}
\end{algorithm}

% Algorithm ~\ref{al:js0} is an approximation for the curvature $\kappa$ with three adjacent points in the path, 
Algorithm~\ref{al:js1} presents a nominated procedure for dynamically selecting the look-ahead configuration. Here, $k_v$ and $k_c$ are objective weights for the velocity and curvature factors. The curvature at the $x$-th configuration is approximated using its two adjacent points. The effect of the curvature $\kappa$ on the look-ahead step can be represented by the decreasing linear function $k_c\kappa + s_{base}$ where $s_{base}$ is the maximum step number corresponding to $\kappa = 0$ (straight segment). If the minimum step number is 0, the coefficient $k_c$ will be determined by $-s_{base} / \pi$.
The minimum look-ahead step $s_{min}$ ensures that the manipulator always progresses forward, even when stationary or on highly curved sections of the path, while the maximum step $s_{max}$ ensures the path will not be heavily overlooked in the opposite cases. To obtain the closest configuration in $\mathbf{Q}$ to $\mathbf{q}$, which is accomplished by the $\mathrm{get\_nearest\_config}$ function, various common search algorithms can be employed, such as linear search, k-nearest neighbours (k-NN), binary search, and KD-trees. The selection of the most appropriate search algorithm depends on several factors, such as the size of the path, the dimensionality of the configuration space, and the desired trade-off between computational efficiency and accuracy. In this study, which focuses on fully actuated 6 DOF and redundant manipulators, the global path is expected to have a high resolution, as depicted in Fig.~\ref{p0}. This implies a large number of steps with small distances between adjacent configurations. Thus, KD-tree can be a suitable implementation for the $\mathrm{get\_nearest\_config}$ function due to its efficiency in handling large, high-dimensional paths. 

The concept of adaptively tracking a predefined path in dynamic environments shares similarities with the approach proposed by Covic et al. \cite{covic-2021}, who addressed the problem of path following when the manipulator deviates from its original trajectory due to obstacle avoidance maneuvers. Their method employs distance information to guide the manipulator back to the reference path by selecting appropriate returning points based on Euclidean distance metrics and path curvature. Unlike their approach, which primarily uses Euclidean distances to determine the nearest point on the global path, this dynamic look-ahead selection incorporates both the manipulator's current joint velocity magnitude and the local path curvature. This dual consideration provides more nuanced tracking behaviour: when moving through straight segments at high velocity, the controller can look further ahead for smoother motion, while in complex curved segments or at lower velocities, it maintains precision by looking closer ahead. Additionally, this approach operates directly in joint space rather than Cartesian space, avoiding the potentially computationally expensive mapping between configuration spaces.

One thing to note here is that the problem's goal does not actually concern tracking the global path geometrically (since this is not possible when obstacles move and block the global path), but only considers the global path as a "guide" to ensure the manipulator can get to the destination. 
% Algorithm ~\ref{al:js2} is a simple implementation for this path-tracking task.
% By eliminating from the search space configurations whose index is no greater than the index of the previous look-ahead configuration ($\mathbf{Q}_{rest} \leftarrow \mathbf{Q} \setminus \mathbf{Q}_{past}$), the above algorithm ensures an 'always forward' behaviour for the manipulator despite being pushed away from the global path due to avoid-obstacle movement.

% \begin{algorithm}
% \caption{Global Path Tracking}
% \label{al:js2}
% \textbf{Input}:  $\mathbf{Q}$, $\mathbf{K}_P$, $\mathbf{K}_D$,  $\Delta t$
% \begin{algorithmic}[1]
% \STATE $\mathbf{q}_e^{prev} \leftarrow \mathbf{0}$
% \WHILE{\textbf{not} $reach\_goal$}
%     \STATE $\mathbf{q}_{next} \leftarrow \mathrm{get\_next\_config}(\mathbf{q}, \mathbf{Q}, {*other\_arguments})$
%     \STATE $\mathbf{q}_e \leftarrow \mathbf{q}_{next}  - \mathbf{q}$
%     \STATE $\mathbf{\dot{q}}_e \leftarrow (\mathbf{q}_e - \mathbf{q}_e^{prev})/ \Delta t$
%     \STATE $\mathbf{\dot{q}} = \mathbf{K}_P\mathbf{q}_e + \mathbf{K}_D\mathbf{\dot{q}}_e$
%     \STATE $\mathbf{q}_e^{prev} \leftarrow \mathbf{q}_e$
% \ENDWHILE
% \end{algorithmic}
% \end{algorithm}

\subsection{Construction of Task-Space Velocity Functions}
\label{sec:VPF}
\subsubsection{Attractive Velocity Function}
\label{sec:Attractive_Velocity_Function}
The attractive velocity function is the task-space velocity function to attract the end effector from its current pose to a goal pose. The corresponding end effector pose can be obtained by applying the forward kinematics for an input joint configuration:
\begin{equation}
\label{avf1}
    \boldsymbol{\xi}_{EE} = \mathrm{fkine}(\mathbf{q}) = 
    \begin{bmatrix}
        \mathbf{R}_{3 \times 3} & \mathbf{t}_{3 \times 1} \\
        0_{1 \times 3} & 1 \\
    \end{bmatrix}
\end{equation}
where $\mathrm{fkine}(\mathbf{q})$ is the forward kinematic solution for the robot end effector, pose as a homogeneous transformation $4 \times 4$ including a rotation $\mathbf{R}_{3 \times 3}$ and a translation $\mathbf{t}_{3 \times 1}$. 

If the robot end effector's current pose is $\mathrm{\boldsymbol{\xi}}_{EE}$ and the goal pose is $\boldsymbol{\xi}_{EE}^{next}$, the error transformation in the current end effector's coordinate frame is:
\begin{equation}
    \label{avf2}
    \boldsymbol{\xi}_{e} = \boldsymbol{\xi}_{EE}\oplus{\boldsymbol{\xi}_{EE}^{next}}^{-1}
\end{equation}

From \eqref{avf2}, the spatial error can be extracted, which is a vector $\mathbf{e}$ of $[e_x, e_y, e_z, e_\alpha, e_\beta, e_\gamma]^T$, where $[e_x, e_y, e_z]^T$ is the translational error and $[e_\alpha, e_\beta, e_\gamma]^T$ is the angular error as the Cardan angles. The potential field at a given pose to attract the robot end effector towards the goal pose can be expressed as:
\begin{equation}
    \label{avf21}
    U_{att}(\mathbf{e}) = \frac{1}{2}K_{att}\|\mathbf{e}\|^2
\end{equation}
where $K_{att}$ is a scalar attractive factor. The attractive velocity of the end effector is to guide the robot in the gradient descent direction of this field, which is:
\begin{equation}
    \label{avf3}
    \mathbf{v}_{att} = -\nabla U_{att} = -K_{att} \mathbf{e}
\end{equation}

From \eqref{avf3}, it can be seen that the attractive velocity required $\mathbf{v}_{att}$ as a spatial vector of $[v_x, v_y, v_z, \omega_x, \omega_y, \omega_z]^T$ can be calculated with a simple proportional control with gain $K_{att}$. 

\subsubsection{Repulsive Velocity Function}
\label{sec:Repulsive_Velocity_Function}
\begin{figure}%[htbp]
\centerline{\includegraphics[width=0.45\linewidth]{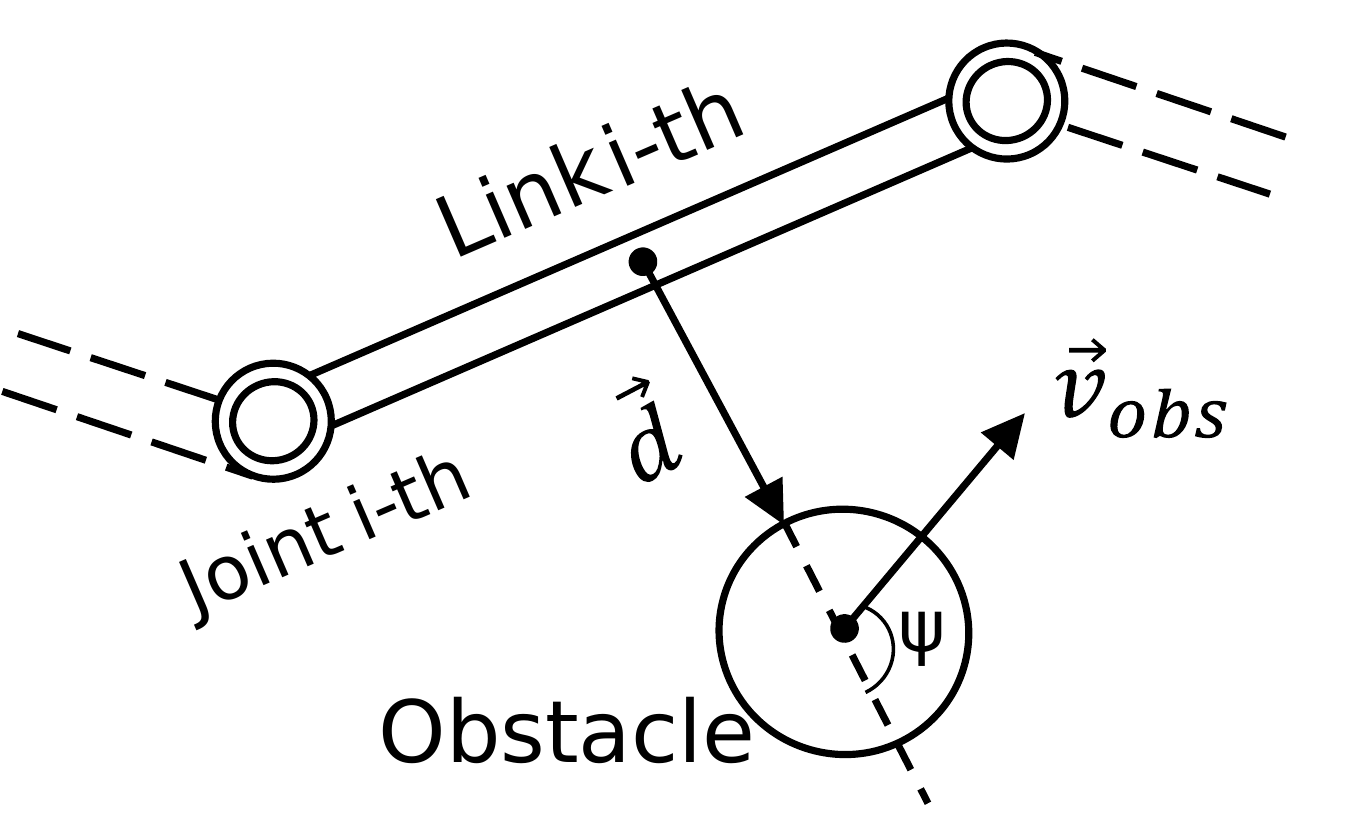}}
\caption{A single manipulator's link in a scenario with a moving obstacle}
\label{p1}
\end{figure}

The repulsive velocity is the task-space velocity used to steer the manipulator away from obstacles. Fig.~\ref{p1} shows the scenario of a single link $i$ -th and its closest obstacle moving with velocity $\mathbf{v}_{obs}$ with respect to the link. The vector $\mathbf{d}$ is the vector connecting the two closest points from the link to the obstacle, and $\psi$ is the angle between $\mathbf{d}$ and $\mathbf{v}_{obs}$, where $0^\circ \leq \psi \leq {180^\circ} $.
 
Similar to the attractive velocity, the repulsive velocity $\mathbf{v}_{rep}$ is demonstrated by a vector of translational and angular components as $[v_{rx}, v_{ry}, v_{rz},\omega_{rx},\omega_{ry},\omega_{rz}]^T$. For the sake of obstacle avoidance, it can be simplified by leaving the angular velocity component to be a zero vector (i.e. $[\omega_{rx},\omega_{ry},\omega_{rz}] = [0,0,0]$) and only working on the translational component. Therefore, the notation $\mathbf{v}_{rep}$ in this section is conveniently considered as the translational repulsive velocity.

In traditional potential field methods for robot navigation, the repulsive potential field $U_{rep}$ is often defined as a function of distance vector $\mathbf{d}$:
\begin{equation}
    \label{rvf1}
    U_{rep}(\mathbf{d}) = 
    \begin{cases}
        0 \quad,\mathrm{if} \textbf{ }\|\mathbf{d}\| \geq d_{max} \\
        \frac{1}{2}K_{rep}(\frac{1}{\|\mathbf{d}\|} - \frac{1}{d_{max}})^2  \quad, \mathrm{otherwise}
    \end{cases}
\end{equation}
where $K_{rep}$ is the repulsive gain and $d_{max}$ is the maximum effect range of the repulsive field. 

The repulsive velocity, which is the negative gradient of the repulsive potential field, can be computed as:
\begin{equation}
    \label{rvf12}
    \mathrm{v}_{rep} = -\nabla U_{rep} =
    \begin{cases}
        0 \quad,\mathrm{if} \textbf{ }\|\mathbf{d}\| \geq d_{max} \\
        -K_{rep}(\frac{1}{\|\mathbf{d}\|} - \frac{1}{d_{max}})\frac{\mathbf{d}}{\|\mathbf{d}\|^2} \quad,\mathrm{otherwise}
    \end{cases}
\end{equation}

The traditional repulsive velocity in \eqref{rvf12} has several limitations. It creates a discontinuity in the robot’s motion when the distance from an obstacle is around the threshold $d_{max}$. This can lead to unpredictable and sudden movements near the effective border of the repulsive field, which is not ideal for a robot system that is expected to operate smoothly. In addition, when an obstacle gets too close to the robot (i.e. $\|\mathbf{d}\|$ is too small), the field calls for an extremely high repulsive velocity without a maximum limit, which may cause undesirable behaviours. Also, \eqref{rvf12} produces a uniform repulsive effect for obstacles at the same distances from the robot. This does not consider the speed and direction of the obstacles’ movement, regardless of whether they are moving slowly or quickly, or moving towards or away from the robot.

\begin{figure}%[htbp]
\centerline{\includegraphics[width=1\linewidth]{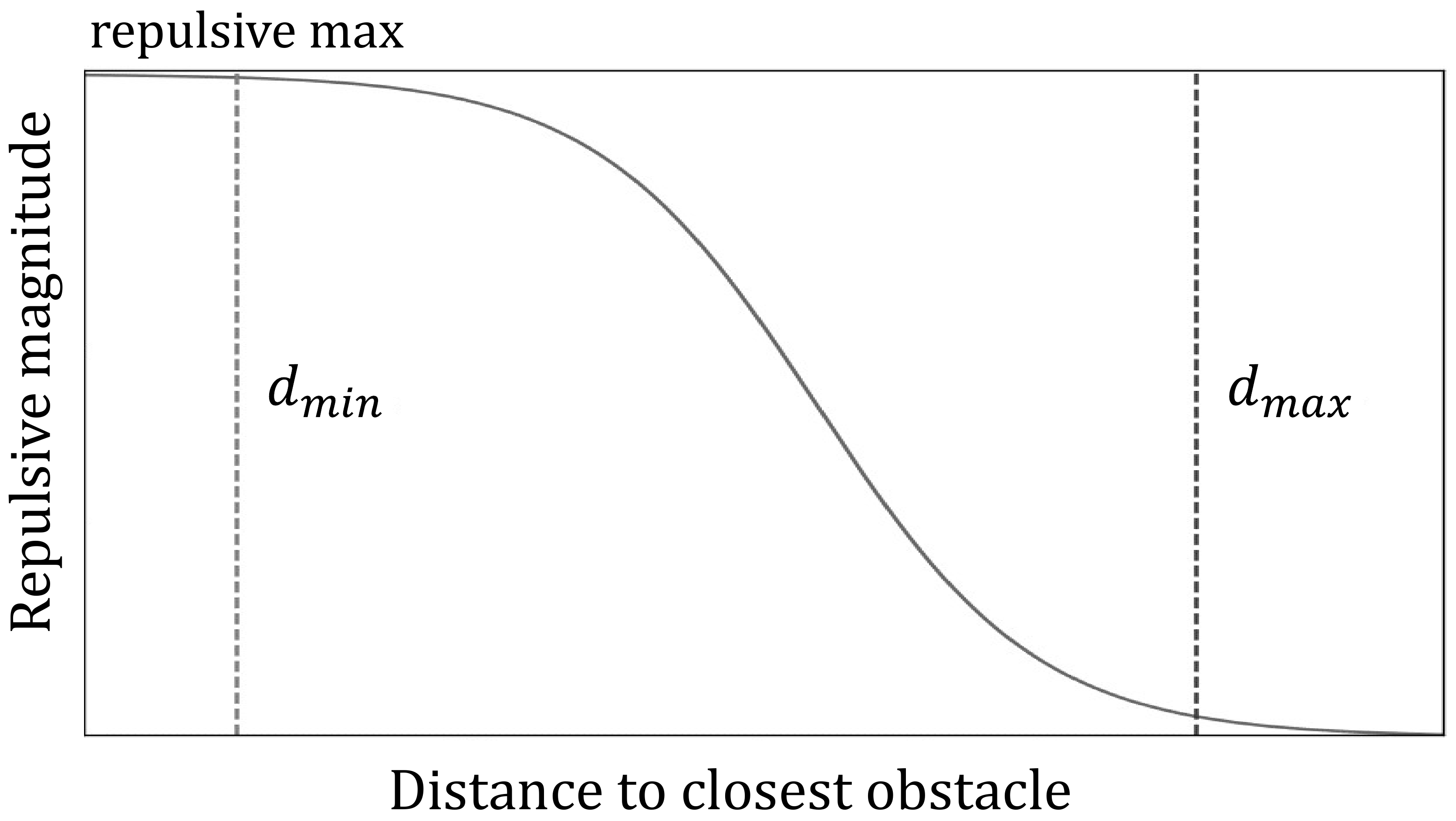}}
\caption{Expected behaviour of the magnitude of the repulsive velocity applied to a link based on the distance to its closest obstacle}
\label{pref11}
\end{figure}

Given $d_{max}$ is the maximum effect range of the repulsive field and $d_{min}$ is the threshold for the smallest distance between a link and its nearest obstacle, it is desirable that this velocity magnitude quickly decays to zero when $\|\mathbf{d}\|$ exceeds $d_{max}$ and quickly saturate to a constant value when $\|\mathbf{d}\|$ is smaller than $d_{min}$, both with a smooth transition. Functions like $\mathrm{sigmoid}$ or $\mathrm{tanh}$ are ideal due to their bounded range and differentiability within their domain. Here, $\mathrm{sigmoid}$ is chosen over the $\mathrm{tanh}$ to construct the repulsive magnitude since this function has a positive range. Thus, no shift along the y-axis is needed. Fig.~\ref{pref11} shows the expected behaviour of the repulsive velocity's magnitude, which has a decreasing sigmoid form. The repulsive velocity can be described by:
\begin{equation}
    \label{rvf2}
    \mathbf{v}_{rep}^{\ i} = \frac{K_{rep}\mathbf{u}}{1 + e^{\alpha d_{max} (\|\mathbf{d}\| - \beta d_{min})}} 
\end{equation}
where $K_{rep}$ is the positive scalar repulsive gain that acts as the upper bound magnitude for the repulsive velocity, and $\alpha$, $\beta$ are hyperparameters to adjust the converging speed of the magnitude of the velocity when the distance $\|\mathbf{d}\|$ approaches $d_{min}$ and $d_{max}$. Here, the graph form of the magnitude of \eqref{rvf2} is similar to that of \eqref{rvf12} within the active region from $d_{min}$ to $d_{max}$, but \eqref{rvf2} has a smooth transition at $\|\mathbf{d}\| =  d_{max}$ and a magnitude cap $K_{rep}$ at $\|\mathbf{d}\| = d_{min}$. Vector $\mathbf{u}$ is a unit vector obtained by $\mathbf{u} = -\mathrm{unit}(\mathbf{d})$, indicates the direction of the acting vector on the link, where the 'minus' sign indicates the effect of pushing the link away from the obstacle.

In case the obstacle position has considerable changes relative to the link (e.g. $\mathbf{v}_{obs}$ has immense magnitude or leads to an abrupt change in the magnitude/direction of $\mathbf{d}$), the strength and direction of the repulsive velocity may need to adapt to the way of which the obstacle is moving relatively to the link for a more efficient avoidance scheme. This adaptability is currently missing from both \eqref{rvf12} and \eqref{rvf2} where only the distance to obstacle term $\mathbf{d}$ is taken into account. Here, the relative velocity $\mathbf{v}_{obs}$ of the obstacle with respect to the link can provide important information about how the distance between the link and the obstacle changes over time and can be decomposed into two components.

% \begin{itemize}
%     \item  
    Firstly, the component of $\mathbf{v}_{obs}$ along the vector connecting the link and the obstacle represents the rate of change in the distance between the obstacle and the robot link. This component has a magnitude that can be calculated using the dot product between $\mathbf{v}_{obs}$ and $\mathbf{u}$, and has $\mathbf{u}$ as its direction:
    \begin{equation}
        \label{rvf200}
        \mathbf{v}_{o,\parallel} = (\mathbf{v}_{obs} \cdot \mathbf{u})\mathbf{u} = -\|\mathbf{v}_{obs}\|\cos({\psi})\mathbf{u}
    \end{equation}
    %
    % \item 
    
    Secondly, the cross product of the obstacle velocity, $\mathbf{v}_{obs}$ and the unit vector, $\mathbf{u}$ results in a perpendicular vector whose magnitude indicates the rate of change in the obstacle movement direction relative to the robot link:
    \begin{equation}
        \label{rvf201}
        \mathbf{v}_{o,\perp} = \mathbf{v}_{obs} \times \mathbf{u} = \|\mathbf{v}_{obs}\|\sin({\psi})\mathbf{n}
    \end{equation}
    where $\mathbf{n}$ is the unit vector normal to the plane of $\mathbf{v}_{obs}$ and $\mathbf{u}$, typically orthogonal to the obstacle's approach direction.
% \end{itemize}

Recognising that \eqref{rvf200} can be directly incorporated into \eqref{rvf2} because it has a similar form of a scalar applied to the vector $\mathbf{u}$. Therefore, the repulsive gain $K_{rep}$ in \eqref{rvf2} can become:
\begin{equation}
    \label{rvf21}
    K_{rep,\parallel} = K_{rep,0} + K_{rep,1}\mathrm{tanh}(\gamma_1 \|\mathbf{v}_{o,\parallel}\|) 
\end{equation}
where the use of the $\mathrm{tanh}$ function is similar to that of the sigmoid in \eqref{rvf2}, which is to smoothly set up the upper and lower bounds for the corresponding repulsive term, and the coefficient $\gamma_1$ is for adjusting the active region. 

Another reason for choosing $\mathrm{tanh}$ over $\mathrm{sigmoid}$ in \eqref{rvf21} is that the $\mathrm{tanh}$ function has its range symmetrical about the horizontal axis, and thus can produce both positive and negative output. In case the second term in \eqref{rvf21} is negative, it means the obstacle's movement tendency is away from the robot link, thus, will result in a weaker repulsive effect, which can reduce the excessive movement for obstacle avoidance and improve the concentration on the main task. Since the value of the second term can vary from $-K_{rep,1}$ to $K_{rep,1}$, the gain parameters can be chosen so that $K_{rep,0} > K_{rep,1} > 0$, which ensures that the repulsive velocity always has $\mathbf{u}$ as its direction; therefore, there will be no pull-toward-obstacle effect when the obstacle suddenly moves away from the robot.    

\begin{figure}%[htbp]
\centerline{\includegraphics[width=\linewidth]{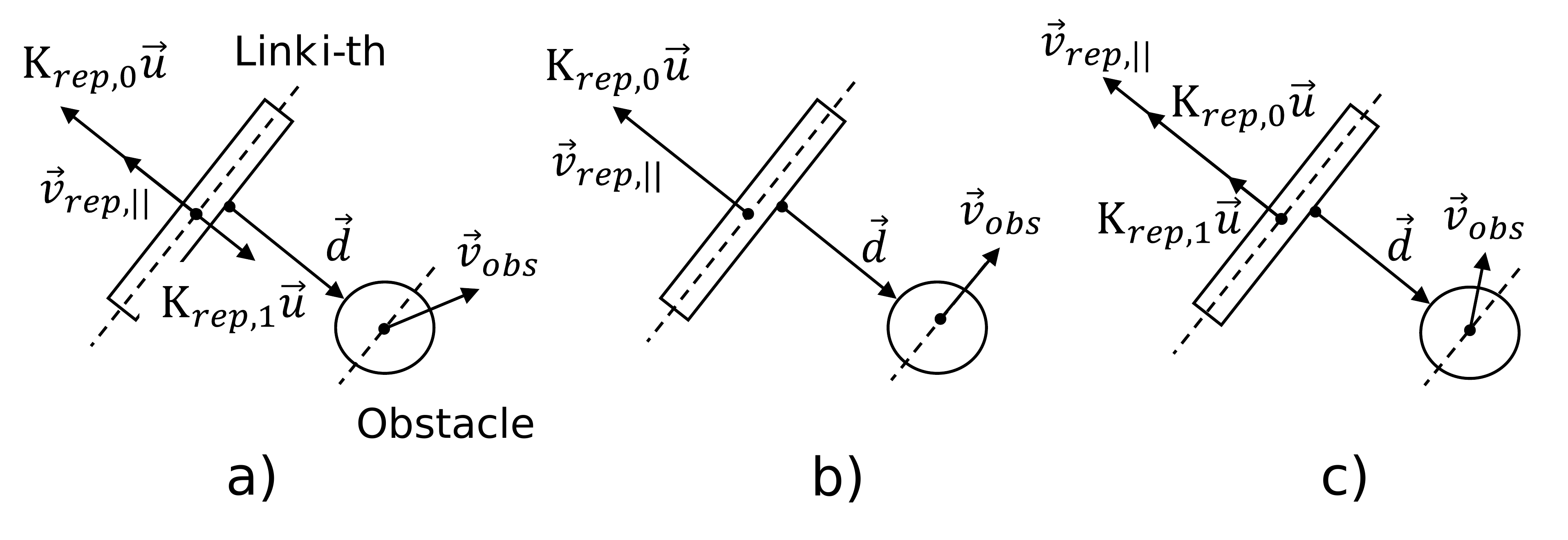}}
\caption{Repulsive velocity in different scenarios: a) Obstacle moves away from the link; b) Obstacle moves along the link; c) Obstacle moves towards the link. Since $\mathbf{d}$ is kept unchanged, the denominator in \eqref{rvf2} can be simplified to 1 for illustration purposes without the loss of generality.}
\label{p22}
\end{figure}

Fig.~\ref{p22} demonstrates the repulsive velocity by \eqref{rvf21} in three scenarios where the obstacle has the same distance to the link but different moving tendencies. Here, the distance vector $\mathbf{d}$ is kept unchanged, which means the traditional repulsive function \eqref{rvf12} will produce identical output repulsive velocity for all cases. By taking into account the obstacle's moving tendency, \eqref{rvf21} performs a better adaptive approach, where the output repulsive velocity $\mathbf{v}_{rep,\parallel}$ can be weakened or strengthened depending on whether the obstacle is moving away or towards the robot link, even if the distance condition is identical. 

The rate of change in the obstacle moving direction, $\mathbf{v}_{o,\perp}$, can also be fed into a $\mathrm{tanh}$ function similar to the second term of \eqref{rvf21} so that its upper and lower bound magnitudes can be controlled via a gain parameter:
\begin{equation}
    \label{rvf22}
    K_{rep,\perp} = K_{rep,2}\mathrm{tanh}(\gamma_2 \|\mathbf{v}_{o,\perp}\|)
\end{equation}

From \eqref{rvf2} to \eqref{rvf22}, the total translational repulsive velocity for the link $i$-th is:
\begin{equation}
    \label{rvf4}
    \mathbf{v}_{rep}^{\ i} = \mathbf{v}_{rep,\parallel}^{\ i} + \mathbf{v}_{rep,\perp}^{\ i} 
\end{equation}
\begin{equation*}
    = \frac{K_{rep,\parallel}}{1 + e^{\alpha d_{max} (\|\mathbf{d}\| - \beta d_{min})}}\mathbf{u} + K_{rep,\perp}\mathbf{n} 
\end{equation*}

A remark here is that the perpendicular term $\mathbf{v}_{rep,\perp}^{\ i}$ does not have an obvious role as the parallel term $\mathbf{v}_{rep,\parallel}^{\ i}$. By producing a movement perpendicular to both the movement direction of the obstacle and the line connecting the link to the obstacle, $\mathbf{v}_{rep,\perp}^{\ i}$ helps reduce the likelihood of collisions by moving the robot link away from the current path of the obstacle. However, by directly producing a repellent effect away from the obstacle, $\mathbf{v}_{rep,\parallel}^{\ i}$ still guarantees a more intuitive and apparent avoidance ability. Therefore, the perpendicular term can be ignored or set to have a much smaller effect than the parallel one. Besides, it is simple to recognise that the magnitude of the repulsive velocity has a bounded maximum value defined by:
\begin{equation}
    \label{rvf40}
    K_{rep,max} = \sqrt{(K_{rep,0} + K_{rep,1})^2 + K_{rep,2}^2} 
\end{equation}
which means the velocity command sent to the robot will always be upper-bounded, thus, ensuring more stability.

If the parameters for computing the repulsive velocity of a link are collected locally to that link, the velocity vector in \eqref{rvf4} is first presented in the coordinate frame of link $i$-th. Given the rotation matrix $\mathbf{R}_{EE}^{\ i}$ from the coordinate frame of the end effector to the coordinate frame of link $i$-th, we can rotate the velocity vector of link $i$-th to be presented in the coordinate frame of the end effector:
\begin{equation}
    \label{rvf41}
    \mathbf{v}_{rep_{EE}}^{\ i} =  \mathbf{R}_{EE}^{\ i} \mathbf{v}_{rep}^{\ i}
\end{equation}
where $\mathbf{R}_{EE}^{\ i}$ is the rotation component of the relative pose $\boldsymbol{\xi}_{EE}^{\ i}$ obtained by inverting the forward kinematic solution from link $i$-th to the end effector. 

It is worth noticing that the closer a link is to the end effector, the more degrees of freedom it has for controllability. Therefore, repulsive commands of the links closer to the end effector should be treated more significantly than those closer to the base. If the manipulator's links are numbered from its base link and $w_i$ is the scalar weight for link $i$-th, then $w_i$ is in increasing order from 1 to N. The repulsive velocity presented in the end effector's coordinate frame to guide the end effector is the weighted sum of the repulsive velocity of each link:
\begin{equation}
    \label{rvf5}
    \mathbf{v}_{rep} = \mathbf{v}_{rep_{EE}} = \sum_{i=1}^{n} w_i\mathbf{v}_{rep_{EE}}^{\ i} \quad, \sum_{i=1}^{n} w_i = 1
\end{equation}
where $n$ is the number of links in the manipulator.     
\subsubsection{Singularity Avoidance in Task-space Control}
\label{sec:Singularity_Avoidance}
Given $\mathbf{J}$ is the Jacobian matrix at the joint configuration $\mathbf{q}$ and $\mathbf{v}$ is the required task-space velocity, the mapping between task-space velocity $\mathbf{v}$ and joint velocity $\mathbf{\dot{q}}$ is:
\begin{equation}
    \label{dls1}
    \mathbf{\dot{q}} = \mathbf{J}^{\dagger} \mathbf{v}
\end{equation}

As seen in \eqref{dls1}, a critical issue in controlling the manipulator with the task-space velocity is the risk of entering a singularity due to the inversion of a singular Jacobian matrix. A common measurement for how far the manipulator is away from a singular configuration is the Yoshikawa number:
\begin{equation}
    \label{dls11}
    \mu(\mathbf{q}) = \sqrt{\mathrm{det}(\mathbf{J}\mathbf{J}^T)}
\end{equation}
  
A common way to deal with singularity is to apply the Damped Least Squares (DLS) method, which trades off the accuracy of the Jacobian matrix inversion process (in other words, the task-space movement), to keep the manipulator away from a configuration that will likely encounter a singularity. Instead of directly pseudo-inverting the Jacobian matrix, \eqref{dls1} can be replaced by:
\begin{equation}
    \label{dls2}
    \mathbf{\dot{q}} = \mathbf{J}^T(\mathbf{J}\mathbf{J}^T + \lambda \mathbf{I}) \mathbf{v}
\end{equation}
where $\lambda$ is the damping factor and $\mathbf{I}$ is the identity matrix. Here, \eqref{dls2} is, in fact, the solution of the objective function $\|\mathbf{J}\mathbf{\dot{q}} - \mathbf{v}\|^2 + \lambda\mathbf{\dot{q}}^2$ without explicit constraints.

The damping factor $\lambda$ can be defined as:
\begin{equation}
    \label{dls3}
    \lambda = 
    \begin{cases}
        (1 - (\frac{\mu}{\epsilon})^2)\cdot \lambda_{max} \quad,\mathrm{if} \textbf{ }\mu < \epsilon \\
         0 \quad, \mathrm{otherwise}
    \end{cases}
\end{equation}
where $\epsilon$ is the manipulability threshold to apply DLS and $\lambda_{max}$ is the maximum damping factor. 

From \eqref{dls2} and \eqref{dls3}, when the manipulator gets closer to a singularity configuration (i.e. $\mu$ decreases), the damping factor $\lambda$ will correspondingly increase. Thus, the manipulator can escape from a singularity configuration by sacrificing some accuracy in the task space motion (i.e. $\mathbf{J} \mathbf{\dot{q}} \neq \mathbf{v}$). However, if the manipulator is not close to a singularity configuration, $\lambda$ becomes zero, and \eqref{dls2} becomes \eqref{dls1}. 
\subsubsection{Modified VPF: Adjustment of Velocity in the Direction of Higher Mobility}
\label{sec:Mobility_Adjustment}
DLS for task-space velocity control prevents the manipulator from nearing a singularity by sacrificing task-space movement accuracy. This leads to configurations where the end effector’s trajectory deviates due to the change from task space to joint space control. DLS application, as in \eqref{dls2}, relies on Yoshikawa manipulability, which is proportional to the manipulability ellipsoid’s volume. A reduced volume due to a smaller semiaxis indicates limited mobility along that axis, but movement along other axes with larger semiaxes may still be possible, eliminating the need for DLS. In contrast, the motion may be requested for a low-mobility task-space direction. However, a large semiaxis may lead to a volume where $\mathbf{\mu} > \mathbf{\epsilon}$; thus, DLS is not activated. In scenarios like real-time obstacle avoidance requiring task-space velocity control, it’s beneficial for the manipulator to maintain sufficient manoeuvring direction, ensuring continuous task-space motion while avoiding obstacles and reaching the goal. Therefore, before feeding the input task-space velocity to the DLS, it is desirable to modify it for enhanced mobility. 
\begin{figure}%[htbp]
\centerline{\includegraphics[width=0.80\linewidth]{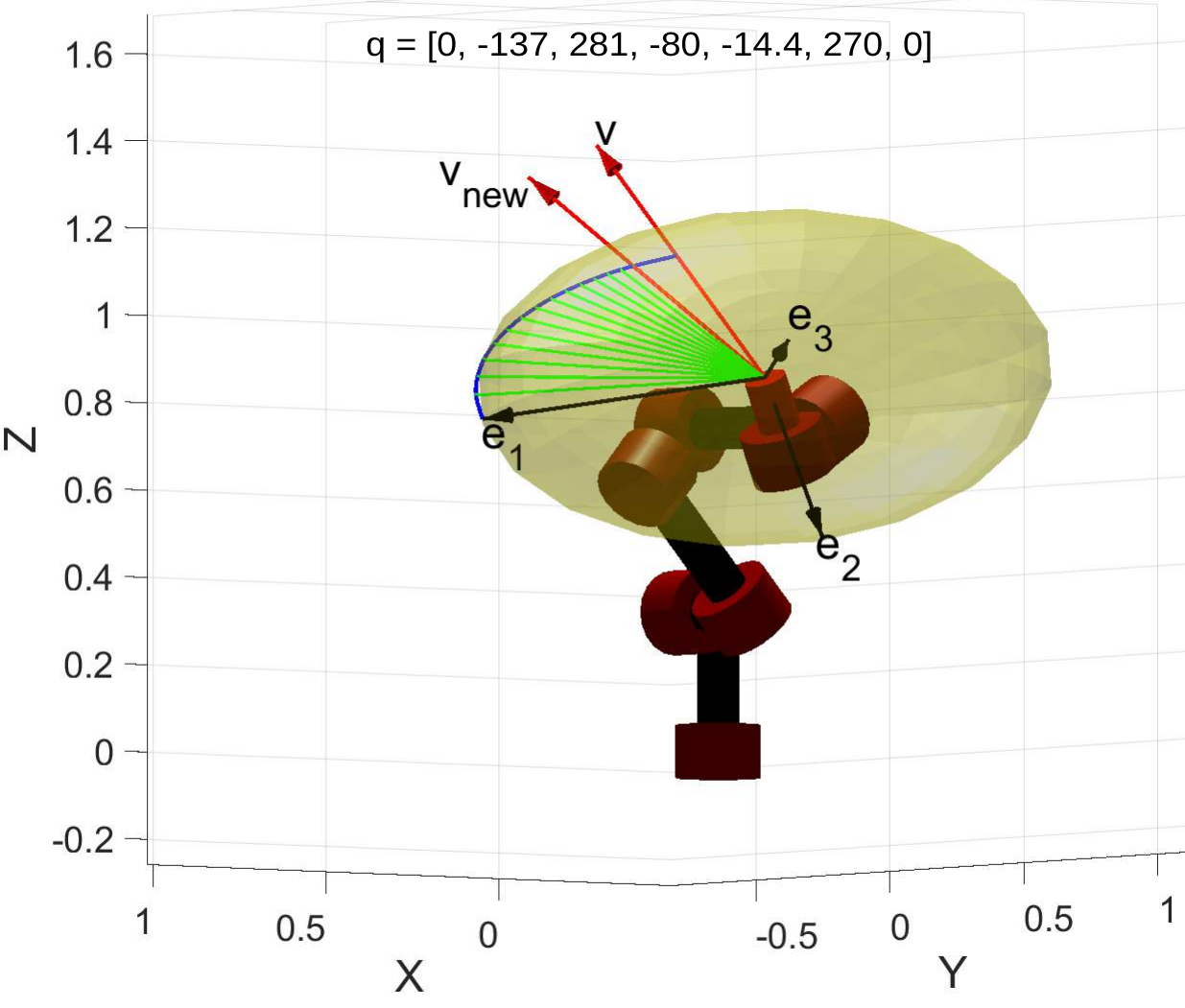}}
\caption{Illustration from a Sawyer Robot with configuration $\mathbf{q}$: the translational manipulability ellipsoid has 3 semiaxes $\mathbf{e}_1$, $\mathbf{e}_2$, $\mathbf{e}_3$. The vector $\mathbf{v}$ is an input translational velocity. The green lines from the ellipsoid's centre to its surface demonstrate how $\mathbf{v}$ intersects the ellipsoid surface when being oriented towards the largest axis $\mathbf{e}_1$, and their lengths indicate the mobility along that direction. Vector $\mathbf{v}_{new}$ is the output translational velocity after orienting $\mathbf{v}$ towards $\mathbf{e}_1$ by a small angle, which has a direction with a higher mobility than the input $\mathbf{v}$.}
\label{pme1}
\end{figure}

The Singular Value Decomposition (SVD) of the Jacobian matrix $\mathbf{J}$, associated with an $n$-DOF manipulator, is articulated as $\mathbf{J} = \mathbf{U} \mathbf{\Sigma} \mathbf{V}^T$, where $\mathbf{J}$ is a $6 \times n$ matrix. Within this decomposition, the columns of $\mathbf{U}$ form an orthonormal basis of dimension $6 \times 6$ that encapsulates the task-space velocity vectors, and $\mathbf{\Sigma}$ is a matrix $6 \times n$ containing the singular values of $\mathbf{J}$. The mobility of the end effector in maneuvering along a specific task-space direction is quantified by the dimensions of a 6-dimensional hyper-ellipsoid having its semiaxes corresponding to the columns of $\mathbf{U}$ and the singular values in $\mathbf{\Sigma}$ representing the magnitude of each semiaxis. If considering only the non-zero submatrix of $\mathbf{\Sigma}$, the hyper-ellipsoid matrix can be defined as $\mathbf{E} = \mathbf{U}\mathbf{\Sigma}$. The capability for the end effector to move along the direction of $\mathbf{v}$ can be examined by positively scaling $\mathbf{v}$ for an intersection with the ellipsoid's surface:
\begin{equation}
    \label{hm2}
    \mu_v(\mathbf{E}, \mathbf{v}) =  \|(\frac{1}{\sqrt{\mathbf{v}^T(\mathbf{E}\mathbf{E}^T)^{-1}\mathbf{v}}})\mathbf{v}\|
\end{equation}
 where $\mathbf{E}$ and $\mathbf{v}$ are represented in the end effector's coordinate frame so that the ellipsoid's centre can be simplified at the origin without losing generality. The result of \eqref{hm2} can be visualised by the length of the green line segments in Fig.~\ref{pme1}. 

Recognise from Fig.~\ref{pme1} that given a fixed configuration ($\mathbf{E}$ is constant), the direction along the largest axis will produce the highest mobility or the global maximum for \eqref{hm2}. Thus, an input task-space velocity can be oriented towards its 'closer' largest semiaxis $\mathbf{e}_{max}$ for a direction with higher mobility ($\mathbf{e}_{max}$ is one of the two largest semiaxes that make with $\mathbf{v}$ an acute angle, i.e. in Fig.~\ref{pme1} if $\mathbf{v}$ points to the opposite direction then $\mathbf{e}_{max} = -\mathbf{e}_1$) using the Rodrigues formula:
\begin{equation}
    \label{hm3}
    \mathbf{v}_{new} = (\mathbf{I} + \mathrm{sin}(\phi_a)\mathbf{K} + (1 - \mathrm{cos}(\phi_a)\mathbf{K}^2)\mathbf{v}
\end{equation}
where $\mathbf{I}$ is the identity matrix, $\phi_a$ is the rotating angle, and $\mathbf{K}$ is the skew-symmetric matrix corresponding to the axis of rotation $\mathbf{k}$, having $\mathbf{k}$ is the unit vector of the cross product between $\mathbf{v}$ and $\mathbf{e}_{max}$. Using the mobility ratio, the condition to decide whether \eqref{hm3} should be applied can be defined as ${\mu_v(\mathbf{E}, \mathbf{v})}/{\|\mathbf{e}_{max}\|} < \zeta$, where $0<\zeta\leq1$.  

It is undesirable if $\mathbf{v}_{new}$ guides the manipulator much closer to the obstacle. This can be considered a double-objective optimisation problem where $\mathbf{v}$ should be guided toward $\mathbf{e}_{max}$ by some angle $\phi_a$, but also the result vector $\mathbf{v}_{new}$ should not be heavily aligned with $\mathbf{d}$, which is the vector from the manipulator to its closest obstacle. Only considering the translational component, the cost function for the rotation angle $\phi_a$ ($0\leq\phi_a\leq\pi/2$) to balance between the alignment with $\mathbf{e}_{max}$ and misalignment with $\mathbf{d}$ can be defined as:
\begin{equation}
\label{hm4}
    \mathrm{f}(\phi_a) = -\omega_1(\mathbf{v}_{new} \cdot \mathbf{e}_{max}) + \omega_2\mathrm{max}(0,(\mathbf{v}_{new} \cdot \mathbf{d}))    
\end{equation}
where $\omega_1$ and $\omega_2$ are objectives weights. If ($\mathbf{v}_{new}\cdot\mathbf{d}) < 0$, then the velocity guides the manipulator away from the obstacle, and the $\mathrm{max}$ function nullifies the second term, which focuses the cost function solely on increasing the mobility. 

\subsubsection{Modified VPF: Countermeasure for Attractive and Repulsive Counterbalance}
The vector sum of the attractive and repulsive velocities specifies the end effector's task-space velocity, $\mathbf{v} = \mathbf{v}_{att} + \mathbf{v}_{rep}$. Stagnation occurs when these velocities counteract each other, leading to a negligible resultant velocity. This happens when their translational components have nearly identical magnitudes but opposite directions.

Without the global information of the environment, one potential solution involves identifying a direction of movement within a plane perpendicular to the axis of the translational attractive or repulsive velocities. Navigating within this plane enables the manipulator to intuitively evade the stagnation state with minimal conflict in adhering to the requirements of the two velocity types, as the direction of movement is temporarily orthogonal to both. Within this perpendicular plane, the axis exhibiting the highest mobility is deemed the direction with the least effort for the manipulator to execute motion along. Therefore, the projection of the manipulability ellipsoid onto this plane can be utilised to identify the direction offering the highest mobility. Given $\mathbf{u}$ is the normalisation of the translational component of $\mathbf{v}_{att}$, the projection matrix onto the plane $\tau$ having $\mathbf{u}$ as its normal vector is $\mathbf{P} = \mathbf{I} - \mathbf{u}  \mathbf{u}^T$, where $\mathbf{I}$ is the identity matrix $3\times3$. With $\mathbf{E}$ as the translational manipulability ellipsoid matrix of the manipulator at the time, the projection of $\mathbf{E}$ onto the plane $\tau$ will be an ellipse described by $\mathbf{E}_{prj} = \mathbf{P}$$\mathbf{E}$. 
% %
% \begin{equation}
%     \label{arc2}
%     \mathbf{E}_{prj} = \mathbf{PE}
% \end{equation}

The axis of greatest mobility aligns with the elongated semiaxis, denoted as $\boldsymbol{\rho}_{max}$, of the ellipse defined by $\mathbf{E}_{prj}$. This semiaxis is intrinsically linked to the column of the highest magnitude. However, considering the inherent symmetry of a semiaxis, two vectors, $\mathbf{p}_1$ and $\mathbf{p}_2$, can satisfy this condition. These vectors display equivalent magnitude but are oriented in diametrically opposite directions. Given the requirement for the manipulator to adhere to a predetermined global trajectory $\mathbf{Q}$, the chosen vector can be the one that guides the manipulator closer towards the next configuration  $\mathbf{q}_{next}$ (from Algorithm \ref{al:js1}) in this global trajectory. With a default speed $v_{def}$, the perpendicular velocity to evade the trapping scenario can be as $\mathbf{v}_{esc} = v_{def} \mathrm{unit}(\boldsymbol{\rho}_{max})$. 

% \begin{algorithm}
% \caption{Select Perpendicular Velocity Direction}
% \textbf{Input}:  $\mathbf{Q}, \mathbf{q}_{curr}, \mathbf{p}_1$, $\mathbf{p}_2$  \\
% \textbf{Output}: $\mathbf{\boldsymbol{\rho}}_{max}$
% \begin{algorithmic}[1]
% \STATE $\mathbf{q}_{next} \leftarrow \mathrm{get\_next\_config}(\mathbf{q}_{curr}, \mathbf{Q})$
% \STATE $\mathbf{p}_{next} \leftarrow \mathrm{get\_position}(\mathrm{fkine}(\mathbf{q}_{next}))$
% \STATE $\mathbf{p}_1, \mathbf{p}_2 \leftarrow \mathrm{predict\_position}(\mathbf{q}_{curr}, \mathbf{p}_1, \mathbf{p}_2, \delta t)$
% \STATE $\mathbf{\boldsymbol{\rho}}_{max} \leftarrow \mathrm{closer\_direction}(\mathbf{p}_{next}, \mathbf{p}_1, \mathbf{p}_2)$
% \STATE \textbf{return} $\mathbf{\boldsymbol{\rho}}_{max}$
% \end{algorithmic}
% \end{algorithm}

\begin{figure}%[htbp]
\centerline{\includegraphics[width=0.85\linewidth]{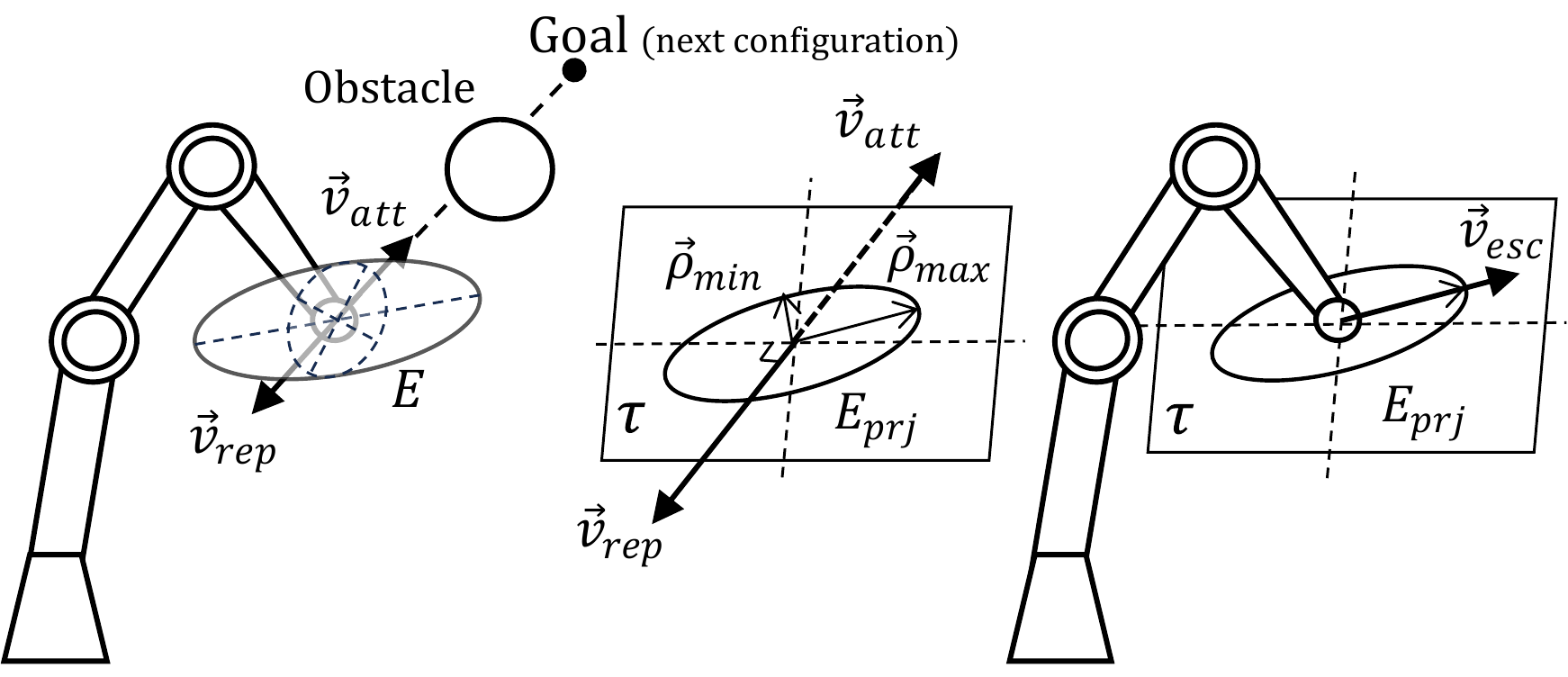}}
\caption{Manipulator escapes a trapping situation by moving along the direction of highest manipulability on the perpendicular plane of the translational attractive and repulsive velocities}
\label{p4}
\end{figure}

Fig.~\ref{p4} demonstrates the process that allows the manipulator to escape stagnation. The plane $\tau$, situated at the end effector and orthogonal to the line of action of the two velocity vectors, hosts the manipulability ellipse resulting from the projection of the manipulability ellipsoid on $\tau$, shaded for emphasis. The semiaxes of the ellipse are denoted by $\mathbf{\boldsymbol{\rho}}_{min}$ and $\mathbf{\boldsymbol{\rho}}_{max}$. 
\subsection{Joint Velocity Command with Constraints}
\label{sec:Command_Constraints}
To solve for the required joint velocity command $\mathbf{\dot{q}}_d$ that satisfies task space velocity $\mathbf{v}$ while respecting joint limits [$\mathbf{q}_{min}$, $\mathbf{q}_{max}$], velocity limits $\mathbf{\dot{q}}_{max}$, and acceleration limits $\mathbf{\ddot{q}}_{max}$, the DLS objective function can be incorporated into a larger quadratic programming (QP) problem:
\begin{equation}
    \label{otm1}
    \mathbf{\dot{q}}_d = \underset{\mathbf{\dot{q}}_d}{\text{argmin}}\|\mathbf{J}\mathbf{\dot{q}}_d - \mathbf{v}\|^2 + \lambda\mathbf{\dot{q}}_d^2
\end{equation}
$\mathrm{s.t:}$
% %
% \begin{equation}
%     \label{otmc1}
%     \frac{\mathbf{q}_{min} - \mathbf{q}}{\Delta t} \leq \mathbf{\dot{q}}_d \leq \frac{\mathbf{q}_{max} - \mathbf{q}}{\Delta t}  
% \end{equation}
% %
% \begin{equation}
%     \label{otmc2}
%     -\mathbf{\dot{q}}_{max} \leq \mathbf{\dot{q}}_d \leq \mathbf{\dot{q}}_{max}
% \end{equation}
% %
% \begin{equation}
%     \label{otmc3}
%      {\mathbf{\dot{q}} - \mathbf{\ddot{q}}_{max}}\Delta t  \leq \mathbf{\dot{q}}_d \leq {\mathbf{\dot{q}} + \mathbf{\ddot{q}}_{max}}\Delta t
% \end{equation}
% %
%
\begin{equation}
    \label{otmc1}
    \frac{q_{min, i} - q_i}{\Delta t} \leq \dot{q}_{d,i} \leq \frac{q_{max, i} - q_{i}}{\Delta t}  \quad, i = 1,2...n
\end{equation}
\begin{equation}
    \label{otmc2}
    |\dot{q}_{d, i}| \leq \dot{q}_{max, i} \quad, i = 1,2...n
\end{equation}
\begin{equation}
    \label{otmc3}
     % \dot{q}_i - \ddot{q}_{max, i}\Delta t  \leq \dot{q}_{d,i} \leq {\dot{q}_i} + \ddot{q}_{max, i}\Delta t
     |\dot{q}_{d,i} - \dot{q}_i| \leq  \ddot{q}_{max, i} \Delta t \quad, i = 1,2...n
\end{equation}
where \eqref{otmc1}, \eqref{otmc2}, \eqref{otmc3} are the joint, velocity and acceleration constraints, respectively. The time step $\Delta t$ is for a control frequency $f = 1/\Delta t$.

For kinematically redundant manipulators or, in general, the cases where the task space's dimension is lower than the joint space's, a non-empty null space $\mathcal{N}(\mathbf{J})$ for the joint space may exist in many configurations. Thus, a more general solution than \eqref{dls1} that makes use of redundancy: 
\begin{equation}
    \label{dls12}
    \mathbf{\dot{q}} = \mathbf{J}^{\dagger} \mathbf{v} + (\mathbf{I} - \mathbf{J}^{\dagger}\mathbf{J})\mathbf{\dot{q}}_0
\end{equation}
where the second term of the right side is the projection onto null space $\mathcal{N}$ of the joint velocity $\mathbf{\dot{q}}_0$ for a secondary objective. The problem \eqref{otm1} can be expanded to be:
\begin{equation}
    \label{otm3}
    \mathbf{\dot{q}}_d = \underset{\mathbf{\dot{q}}_d}{\text{argmin}}\|\mathbf{J}\mathbf{\dot{q}}_d - \mathbf{v}\|^2 + \lambda\mathbf{\dot{q}}_d^2 + \alpha\|(\mathbf{I} - \mathbf{J}^\dagger \mathbf{J})(\mathbf{\dot{q}}_0 - \mathbf{\dot{q}}_d)\|^2
\end{equation}
where $\alpha$ is a weighting factor for the secondary objective. 

Several common approaches for the secondary objectives include joint centering, manipulability maximisation, obstacle avoidance, etc. To enhance the efficiency of the task-space control, $\mathbf{\dot{q}}_0$ can be designed for manipulability improvement:
\begin{equation}
    \label{otm4}
    \mathbf{\dot{q}}_0 = k_m\nabla_{\mathbf{q}}\mu(\mathbf{q})
\end{equation}
where $k_m$ is a positive gain, and $\nabla_{\mathbf{q}}\mu(\mathbf{q})$ is the gradient of \eqref{dls11} with respect to the joint state $\mathbf{q}$.

\subsection{Overall System for Path Tracking and Obstacle Avoidance}
The integration of a global path-tracking controller (\ref{sec:Global_Path_Tracking}) with a reactive obstacle avoidance controller (\ref{sec:VPF}) offers a more efficient solution for manipulator navigation in dynamic environments. This dual-control approach maintains focus on the overall efficient path while allowing for reactive behaviour when necessary. The global path-tracking controller ensures convergence to the desired trajectory, efficient recovery after obstacle avoidance maneuvers, and also helps maintain task performance and predictability when obstacles are not present. The system comprises several key components working in tandem to ensure efficient and safe navigation.

A Look-ahead Calculation module (\ref{sec:Global_Path_Tracking}) processes the global path and the current manipulator's state to determine future desired joint configurations. These configurations serve as inputs for two primary control modules: a PD Controller for path tracking/recovery and a Velocity Potential Field (VPF) module for obstacle avoidance. The VPF module incorporates both attractive movement towards the goal, mitigating potential deadlocks, and repulsive movement for obstacle evasion.

The system employs a Selector mechanism to arbitrate between the path-tracking velocity from the PD Controller and the avoidance velocity from the VPF. This mechanism prioritises obstacle avoidance when significant repulsive forces are present. The control input $\mathbf{\dot{q}}$ switches between the path tracking velocity $\mathbf{\dot{q}}_{global}$ (\ref{sec:Global_Path_Tracking}) and the VPF velocity $\mathbf{\dot{q}}_{local}$ (\ref{sec:VPF}) as follows:
\begin{equation}
\label{fs1}
\mathbf{\dot{q}} = (1 - \nu)\mathbf{\dot{q}}_{global} + \nu\mathbf{\dot{q}}_{local}
\end{equation}
where $\nu \in {0, 1}$ is a switching logic based on obstacle proximity. Given $d_{max}$ as the maximum effect range of the repulsive field and $d$ as the distance from the manipulator to the nearest obstacle, $\nu$ is defined as:
\begin{equation}
\label{fs2}
\nu =
\begin{cases}
1 \quad, \mathrm{if} \textbf{ }d < d_{max} \\
0 \quad, \mathrm{otherwise}
\end{cases}
\end{equation}

To ensure safe operation, a Command Calculation block processes the selected velocity, ensuring that the final commands sent to the manipulator adhere to its implicit constraints, as detailed in Section \ref{sec:Command_Constraints}.

This architecture effectively balances proactive path following with reactive obstacle avoidance, providing a comprehensive solution for autonomous navigation in complex, obstacle-cluttered environments. The system's modular design allows for efficient path tracking while maintaining the flexibility to quickly respond to dynamic obstacles, ensuring safe and efficient manipulator operation.

\subsection{Hybrid Planner: Proof of Completeness and Soundness}

This section provides proof of completeness and soundness for the proposed hybrid planner, ensuring it plans and guides the manipulator from an initial to an end pose while avoiding static and dynamic obstacles along the path. All necessary assumptions are outlined below.

\textbf{Assumption 1}
Dynamic obstacles presented within the context of this paper are assumed to be convex or can be approximated as convex.

\textbf{Assumption 2}
Dynamic obstacles presented in this paper are assumed to have their states (i.e. the closest distance to the manipulator and velocity as in \ref{sec:Repulsive_Velocity_Function})
measured at all times.

\textbf{Theorem 1}
An RRT* global planner for a given robot manipulator guarantees a path connecting the initial and end positions of the robot's effector while adhering to kinematic constraints.

\textit{Proof} Proofs for this theorem are outlined in theorem 38 of~\cite{salzman-2016}, which provides the full proof of RRT* optimality. 

Since the global RRT* planner can produce a collision-free (with static obstacles) connected path between two points, it is now up to the local planner to efficiently track the planned path while avoiding dynamic obstacles.

\textbf{Theorem 2}
The proposed hybrid planner, which combines the global RRT* path with the local potential field-based planner, guarantees completeness and soundness for the manipulator's motion planning problem.

\textit{Proof} Consider a desired configuration $\mathbf{q}_d$ on the global path, and $U(\mathbf{q}) \geq 0$ is the potential field in joint space. The error between the current configuration $\mathbf{q}$ and $\mathbf{q}_d$ is $\mathbf{q}_e = \mathbf{q}_d - \mathbf{q}$. 

% \textit{Lemma}: 

The Lyapunov candidate function can be chosen as follows:
\begin{equation}
    \label{lc1}
    V(\mathbf{q}) = \frac{1}{2}\|\mathbf{q}_e\|^2 + U(\mathbf{q})
\end{equation}

Taking time derivative of \eqref{lc1}:
\begin{equation}
    \label{lc2}
    \dot{V}(\mathbf{q}) = \mathbf{q}_e^T\mathbf{\dot{q}} + \nabla{U}(\mathbf{q})^T\mathbf{\dot{q}} 
\end{equation}    

Substitute the control law from \eqref{fs1}:
\begin{equation}
    \label{lc3}
    \dot{V}(\mathbf{q}) = (\mathbf{q}_e + \nabla{U}(\mathbf{q}))^T[(1-\nu)\mathbf{\dot{q}}_{global} + \nu\mathbf{\dot{q}}_{local}]     
\end{equation}    

For path tracking mode ($\nu = 0$), the field effect is negligible:
\begin{equation}
    \label{lc4}
    \dot{V}(\mathbf{q}) = \mathbf{q}_e^T\mathbf{\dot{q}}_{global}     
\end{equation}    

Since $\mathbf{\dot{q}}_{global}$ is the output of a PD controller, it can have the form $-k\mathbf{q}_e$ where $k$ is a positive scalar. Therefore, \eqref{lc4} can be rewritten as: 
\begin{equation}
    \label{lc5}
    \dot{V}(\mathbf{q}) = -k\mathbf{q}_e^T\mathbf{q}_e = -k\|\mathbf{q}_e\|^2   
\end{equation}

For obstacle avoidance mode ($\nu = 1$), \eqref{lc3} becomes:
\begin{equation}
    \label{lc6}
    \dot{V}(\mathbf{q}) = (\mathbf{q}_e + \nabla{U}(\mathbf{q}))^T\mathbf{\dot{q}}_{local}  
\end{equation}

Since the nature of the local velocity is the negative gradient of the potential field, $\mathbf{\dot{q}}_{local} = -\nabla{U}(\mathbf{q})$, \eqref{lc6} becomes:
\begin{equation}
    \label{lc7}
    \dot{V}(\mathbf{q}) = \mathbf{q}_e^T\mathbf{\dot{q}}_{local} - \|\mathbf{\dot{q}}_{local}\|^2  
\end{equation}

The local velocity is the composition of the attractive velocity $\mathbf{\dot{q}}_{att}$ and the repulsive velocity $\mathbf{\dot{q}}_{rep}$. Also, $\mathbf{\dot{q}}_{att} = -K_{att}\mathbf{q}_e$. Therefore, \eqref{lc7} can be rewritten as:
\begin{equation}
    \label{lc8}
     \dot{V}(\mathbf{q}) = \mathbf{q}_e^T\mathbf{\dot{q}}_{rep} -K_{att}\|\mathbf{q}_e\|^2 -\|\mathbf{\dot{q}}_{rep} -K_{att}\mathbf{q}_{e}\|^2  
\end{equation}
% %
% \begin{equation}
%     \label{lc8}
%      \dot{V}(\mathbf{q}) = 
%      (2K_{att}+1)\mathbf{q}_e^T\mathbf{\dot{q}}_{rep} - 2K_{att}|\mathbf{q}_e\|^2 - \|\mathbf{\dot{q}}_{rep}\|^2 
% \end{equation}
It is clear from \eqref{lc5} that in the path-tracking mode where the field effect is negligible, $\dot{V}(\mathbf{q})$ is always negative and only equals zero when the robot reaches its goal. The system in this mode, therefore, is asymptotically stable when $\mathbf{q}_d$ converges to the goal configuration.

For the obstacle avoidance mode, from \eqref{lc8}, the system tends towards stability as long as the negative terms dominate, which is satisfied when the repulsive velocity is not significantly larger than and opposes the attractive velocity. When the manipulator is driven toward a goal obstructed by the obstacle, $\mathbf{q}_e^T\mathbf{\dot{q}}_{rep}$ will grow large for the positive term to dominate, ensuring the Lyapunov function increases, discouraging the manipulator from moving closer. This means that stability will be sacrificed in favor of avoiding collisions, for example, in case the obstacle is too close to the destination making it unreachable. At this time, under the effect of the potential field, the manipulator will hover in space. However, since the magnitude of the repulsive velocity is always upper-bounded (as in \ref{sec:Repulsive_Velocity_Function}), the system will not be pushed uncontrollably away from stability, thus, ensuring its recovery behaviour.   

% -------------------------------------------------

\section{Experiments and Results}
This section presents simulations that evaluate and compare the effectiveness of two planning methods. The first is the traditional method known as the Velocity Potential Function (VPF), and the second is the new hybrid method, which is the primary subject of this study. The comparison involves investigating how the new hybrid method compares with the well-established VPF method in various scenarios. Since the VPF planner is commonly used in unpredictable environments, it provides a useful benchmark for this comparison.

The experimental setup involves the Sawyer robot, a 7-DOF manipulator from Rethink Robotics. Sawyer has a payload of 4 kg with a maximum reach of 1260 mm and is known for its lightweight design and active compliance, making it a popular choice for human-robot interaction tasks and experiments. The simulations were developed using Python 3.10.11, with Robotics Toolbox for Python version 1.1.0 as the primary support package~\cite{corke-2021}. Computations and visualisation are conducted under an Intel(R) Core(TM) i7-10750H CPU and an NVIDIA GeForce RTX 2060 graphics card.

The experiment compares the proposed hybrid planner with the traditional VPF in guiding the Sawyer from an initial to a final configuration in an environment with moving obstacles. The simulation outcomes provide an analytical comparison of their performance, highlighting their strengths and weaknesses in dynamic motion planning. The principal parameters of both planners are listed in Table~\ref{table1}, and validated through experiments to ensure meaningful performance. For the hybrid planner, a basic RRT* algorithm~\cite{salzman-2016} creates a near-optimal path in the joint space, using a cost function to generate a collision-free path. The goal is to minimise travel distance while maintaining high manipulability and a safe distance from obstacles, expressed as:
\begin{equation}
    \label{exp1}
    C = \omega_{p}L + \omega_{m}M^{-1} + \omega_oD^{-1}
\end{equation}
where $\omega_{p}$, $\omega_{m}$, and $\omega_{o}$  are the weighting factors for the path length, manipulability and closest distance to obstacles. $L$ is calculated as the sum of Euclidean distances between consecutive configurations in joint space, while $Manip$ is the manipulability index by Yoshikawa as in \eqref{dls11} and $C$ is the distance from the manipulator to its closest obstacles. Each cost criterion variable ($L$, $M$ and $C$) can be normalised (such as filtered through a sigmoid function to output an identical range from 0 to 1), allowing the weighting factors to accurately adjust the importance of each factor. 
In these experiments, $\omega_o$ can be set to zero for several reasons. Firstly, if the obstacles move unpredictably, the $C$ factor loses its relevance during movement. Secondly, this factor incurs high computational costs for constantly checking the distances from each robot link to each obstacle. Finally, without the distance-to-obstacle factor, the hybrid planner's ability to react to the movement of obstacles can be tested more efficiently.    

% \begin{table}%[htbp]
%   \caption{Experiment Parameters}
%   \centering
%   \label{table1}
%   \begin{adjustbox}{max width =\columnwidth}
%   \begin{tabular}{|l|c|c|l|} \hline  
    
%      \textbf{Section}&\multicolumn{2}{|c|}{\textbf{Hybrid}}&  \textbf{VPF}\\ \hline  
    
%       \ref{sec:Global_Path_Tracking}&$k_v$, $s_{base}$, $s_{min}$, $s_{max}$& $5, 5, 1, 10$  &\\ \hline  
    
%       &$K_P$, $K_D$& $200, 100$  &$K_{att} = 1.5$\\ \hline  
    
%       \ref{sec:Attractive_Velocity_Function}, \ref{sec:Repulsive_Velocity_Function}&$K_{att}$, $K_{rep,0}$, $K_{rep,1}$, $K_{rep,2}$ & $1.5, 0.5, 0.2, 0.1$  &$K_{rep} = 0.5$\\ \hline  
%   &$d_{min}$, $d_{max}$, $\alpha$, $\beta$& $0.01$m, $0.2$m, $200$, $12.5$ &$d_{max} = 0.2m$\\ \hline  
    
%       &$\mathbf{\omega}$& $[0, 0.1, 0.2, 0.4, 0.6, 0.8, 1]/7$  &\\ \hline  
    
%        \ref{sec:Mobility_Adjustment}&$\zeta$, $\omega_1$, $\omega_2$& $0.7, 1, 1$  &\\ \hline  
    
%       &\multicolumn{3}{|c|}{\textbf{Mutual Parameters}}\\ \hline  
    
%       \ref{sec:Singularity_Avoidance}&$\epsilon$, $\lambda$ &  \multicolumn{2}{|c|}{$0.01, 0.5$}\\ \hline  
    
%       \ref{sec:Command_Constraints}&$\mathbf{q}_{max}$, $\mathbf{q}_{min}$ (deg)&  \multicolumn{2}{|c|}{$\pm[170, 120, 170, 120, 170, 120, 175]$}\\ \hline  
    
%       &$\mathbf{\dot{q}}_{max}$, $\mathbf{\ddot{q}}_{max} (\mathrm{deg}/s, \mathrm{deg}/s^2)$&  \multicolumn{2}{|c|}{$35, 70$ per joint}\\\hline 
%   \end{tabular}
%   \end{adjustbox}
% \end{table}

\begin{table}%[htbp]
  \caption{Experiment Parameters}
  \centering
  \label{table1}
  \begin{adjustbox}{max width =\columnwidth}
  \begin{tabular}{|c|c|c|c|} \hline  
    
     \textbf{Section}&\multicolumn{2}{|c|}{\textbf{Hybrid}}&\textbf{VPF}\\ \hline  
    
      % \multirow{2}{*}{\ref{sec:Global_Path_Tracking}}&$k_v$, $s_{base}$, $s_{min}$, $s_{max}$& $5, 5, 1, 10$  &\multirow{6}{*}{$K_{att} = 1.5$, $K_{rep} = 0.5$, $d_{max} = 0.2m$}\\ \cline{2-3}

      \multirow{2}{*}{\ref{sec:Global_Path_Tracking}}&$k_v$, $s_{base}$, $s_{min}$, $s_{max}$& $5, 5, 1, 10$  &\multirow{6}{*}{\makecell{$K_{att} = 1.5$ \\ $K_{rep} = 0.5$ \\$d_{max} = 0.2m$}}\\ \cline{2-3}

      & $K_P$, $K_D$& $200, 100$  & \\ \cline{1-3} %\hline  
    
      \multirow{3}{*}{\ref{sec:Attractive_Velocity_Function}, \ref{sec:Repulsive_Velocity_Function}}&$K_{att}$, $K_{rep,0}$, $K_{rep,1}$, $K_{rep,2}$ & $1.5, 0.5, 0.2, 0.1$  &\\ \cline{2-3}
      &$d_{min}$, $d_{max}$, $\alpha$, $\beta$& $0.01$m, $0.2$m, $200$, $12.5$ &\\ \cline{2-3}  
    &$\mathbf{\omega}$& $[0, 0.1, 0.2, 0.4, 0.6, 0.8, 1]/7$  &\\   \cline{1-3}
    
       \ref{sec:Mobility_Adjustment}&$\zeta$, $\omega_1$, $\omega_2$& $0.7, 1, 1$  &\\ \hline  
    
      \multicolumn{4}{|c|}{\textbf{Mutual Parameters}}\\ \hline  
    
      \ref{sec:Singularity_Avoidance}&$\epsilon$, $\lambda$ &  \multicolumn{2}{|c|}{$0.01, 0.5$}\\ \hline  
  
    \multirow{2}{*}{\ref{sec:Command_Constraints}}&$\mathbf{q}_{max}$, $\mathbf{q}_{min}$ (deg)&  \multicolumn{2}{|c|}{$\pm[170, 120, 170, 120, 170, 120, 175]$}\\ \cline{2-4} 
    &$\mathbf{\dot{q}}_{max}$, $\mathbf{\ddot{q}}_{max} (\mathrm{deg}/s, \mathrm{deg}/s^2)$&  \multicolumn{2}{|c|}{$35, 70$ per joint}\\\hline 
  \end{tabular}
  \end{adjustbox}
\end{table}

% \begin{table}%[htbp]
%   \caption{Parameters for VPF and Hybrid Planners}
%   \centering
%   \label{table1}
%   \begin{adjustbox}{max width =\columnwidth}
%   \begin{tabular}{|c|c|c|c|}
%     \hline
%     \multicolumn{2}{|c|}{\textbf{Parameters}} & \textbf{VPF} & \textbf{Hybrid} \\
%     \hline
%     $K_{att}$ & Gain for attractive velocity & 2 & 6 \\
%     \hline
%     $d_{max}$ & Max. effect range of repulsive field & 0.2m& 0.2m\\
%     \hline
%     $d_{min}$ & Min. distance between link and obstacle & x& 0.01m \\
%     \hline
%     $K_{rep}$ & Gain for repulsive velocity & $1/80$ & $1/80$ \\
%     \hline
%     $\epsilon$ & Manipulability threshold for applying DLS & 0.1 & 0.1 \\
%     \hline
%     $\lambda_{max}$ & DLS max. damping factor & 0.5 & 0.5 \\
%     \hline
%     $\zeta$ &  \begin{tabular}[c]{@{}c@{}} Proportion of the maximum singular \\ value for a direction mobility threshold \end{tabular} & x & 0.8 \\ 
%     \hline
%     $\gamma$ & Adjustment rate for a higher mobility direction & x & 0.2 \\
%     \hline
%     $i_{max}$ & Max. iterations for a higher mobility direction  & x & 5 \\
%     \hline
%     $v_{def}$ & Default magnitude for a perpendicular velocity  & x & 1m/s \\
%     \hline
%     $\Delta \mathbf{q}_{max}$ & \begin{tabular}[c]{@{}c@{}}Max. Euclidean distance between two adjacent\\ configurations within the global path\end{tabular} & x & $5^o$ \\
%     \hline
%     $\Delta t$ & \begin{tabular}[c]{@{}c@{}}  Time step for moving between\\ two adjacent joint configurations \end{tabular}  & 0.05s & 0.05s \\
%     \hline
%   \end{tabular}
%   \end{adjustbox}
% \end{table}

%\FloatBarrier

\subsection{Simulation}
\begin{figure}%[htbp]
\centerline{\includegraphics[width = 0.8\linewidth]{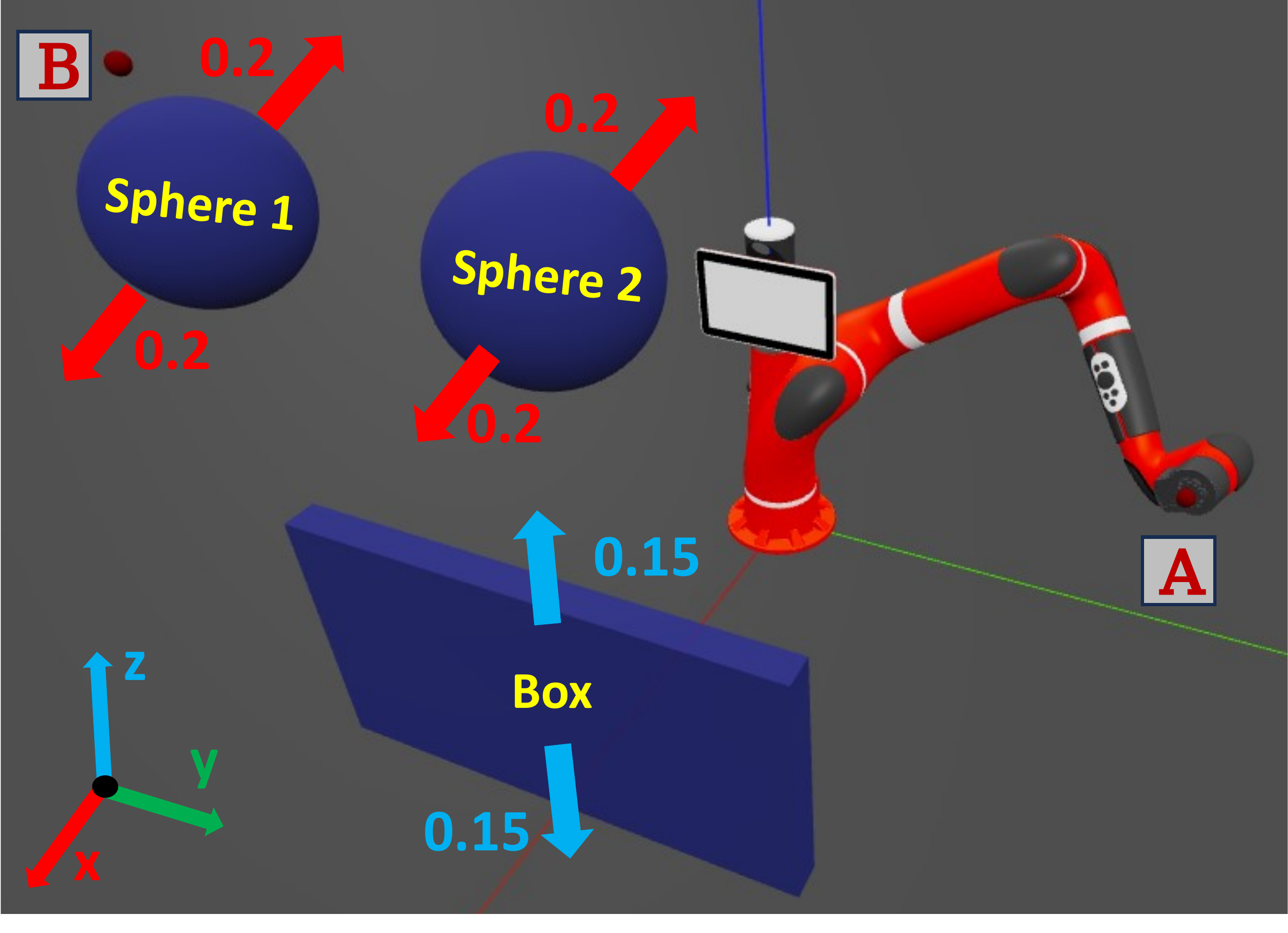}}
\caption{Initial position of the Sawyer with initial end effector's position (A) and goal position (B). Three obstacles are shown with their centroids in the initial position, and their movement ranges and axes annotated.}
\label{p5}
\end{figure}

Fig.~\ref{p5} illustrates the first experimental setup, featuring the manipulator and obstacles. The Sawyer manipulator's base is positioned at the origin, with its homogeneous transform as the identity matrix. The initial and goal configurations ordered from base to end effector are [90, -33, 150, -87, -77, -73, 1] and [-90, -45, 165, 35, 100, -80, 76] degrees, respectively. In Fig.~\ref{p5}, three obstacles fluctuate around their initial positions along given axes. The obstacles' features are shown in Table \ref{table2}. Parameters denoting the size or position of objects conform to the conventional $\mathbb{R}^3$ Cartesian coordinate system and are expressed in standard SI units for clarity and consistency.

\begin{table}%[htbp]
  \centering
  \caption{Obstacles'Features in Simulation Setup}
  \label{table2}
  \begin{adjustbox}{max width =\columnwidth}
  \begin{tabular}{|c|c|c|c|}
    \hline
    \ & \textbf{Sphere 1} & \textbf{Sphere 1} & \textbf{Box} \\
    \hline
    Dimension (m) & radius = 0.1 & radius = 0.1 & [0.05, 0.75, 0.4]\\
    \hline
    Initial Position (m) & [0.8,-0.2,0.9]& [0.8,0.2,0.9]& [0.7,0,0.2]\\
    \hline
    Speed (m/s) & 0.3& 0.1& 0.3\\
    \hline
  \end{tabular}
  \end{adjustbox}
\end{table}

\begin{figure}%[htbp]
\centerline{\includegraphics[width=1\linewidth]{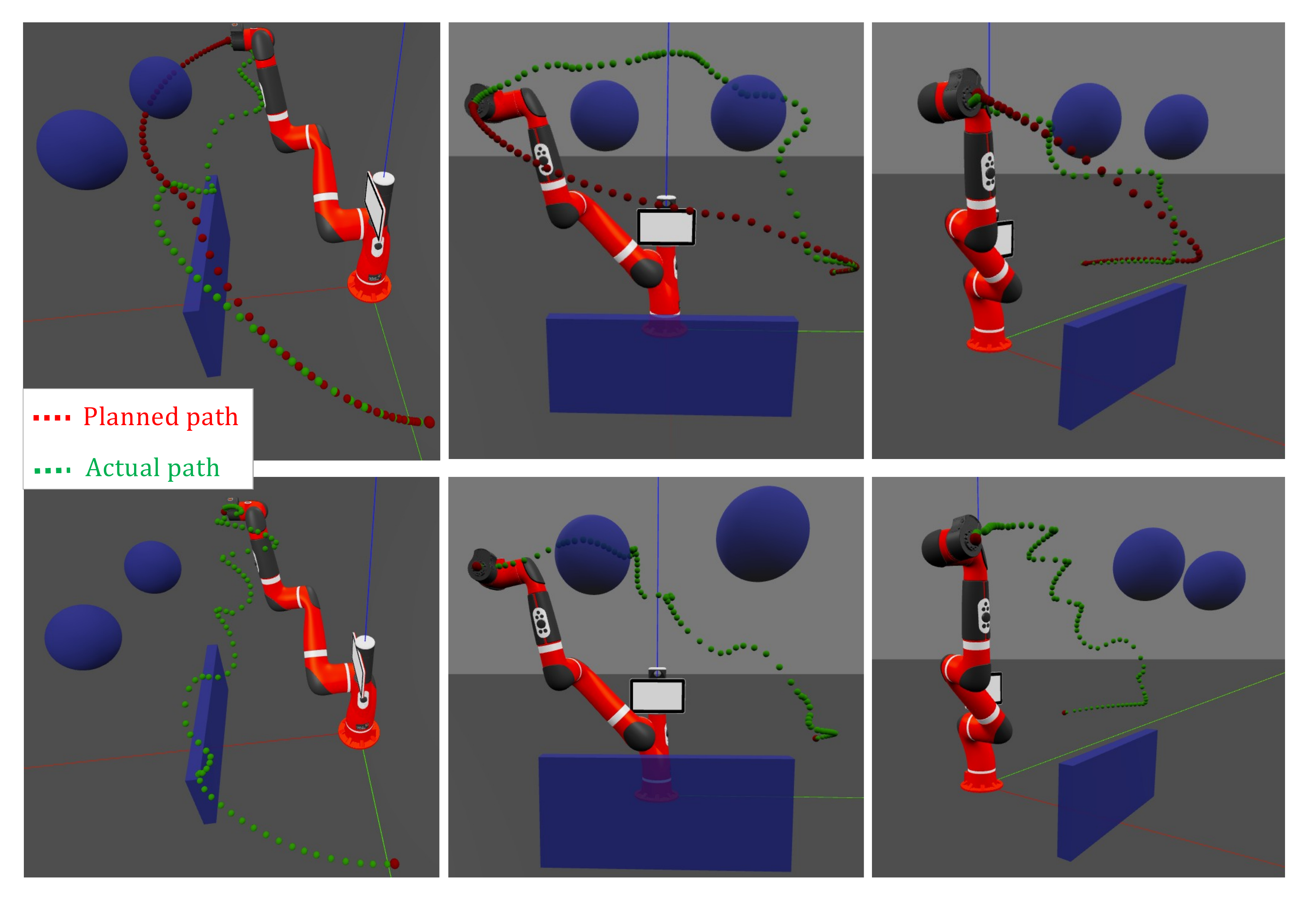}}
\caption{Trajectory from A to B by the end effector's positions for several runs. Top: Hybrid Planner. Bottom: VPF Planner.}
\label{p7}
\end{figure}

To ensure randomness for each test run to observe the planners' reactions to obstacles in unknown configurations, the obstacles will be moved first. The manipulator will then start after a random interval, including the global planning phase for the hybrid planner. Fig.~\ref{p7} provides a comparative visualisation of the hybrid and VPF planners in three test runs, with the former depicted at the top and the latter at the bottom. The green line represents the actual trajectory of the end effector throughout its motion, while the red line represents the initially discovered global path (only the hybrid planner).

In the hybrid planner, the green line's deviation from the red line indicates an adaptive response to changing obstacle positions, balancing collision avoidance with global path tracking. The hybrid planner's path shows somewhat better orderliness, which may explain its ability to maintain a safer distance from obstacles. This distinction arises because the hybrid planner, with global information, only prioritises obstacle avoidance when needed, while the VPF planner constantly adapts to environmental changes en route to the goal. 

\begin{figure*}[t]
    \centering
    \begin{tabular}{@{}c@{}c@{}c@{}c@{}c@{}}
        \scalebox{1.0}[1.1]{\includegraphics[width=0.19\textwidth]{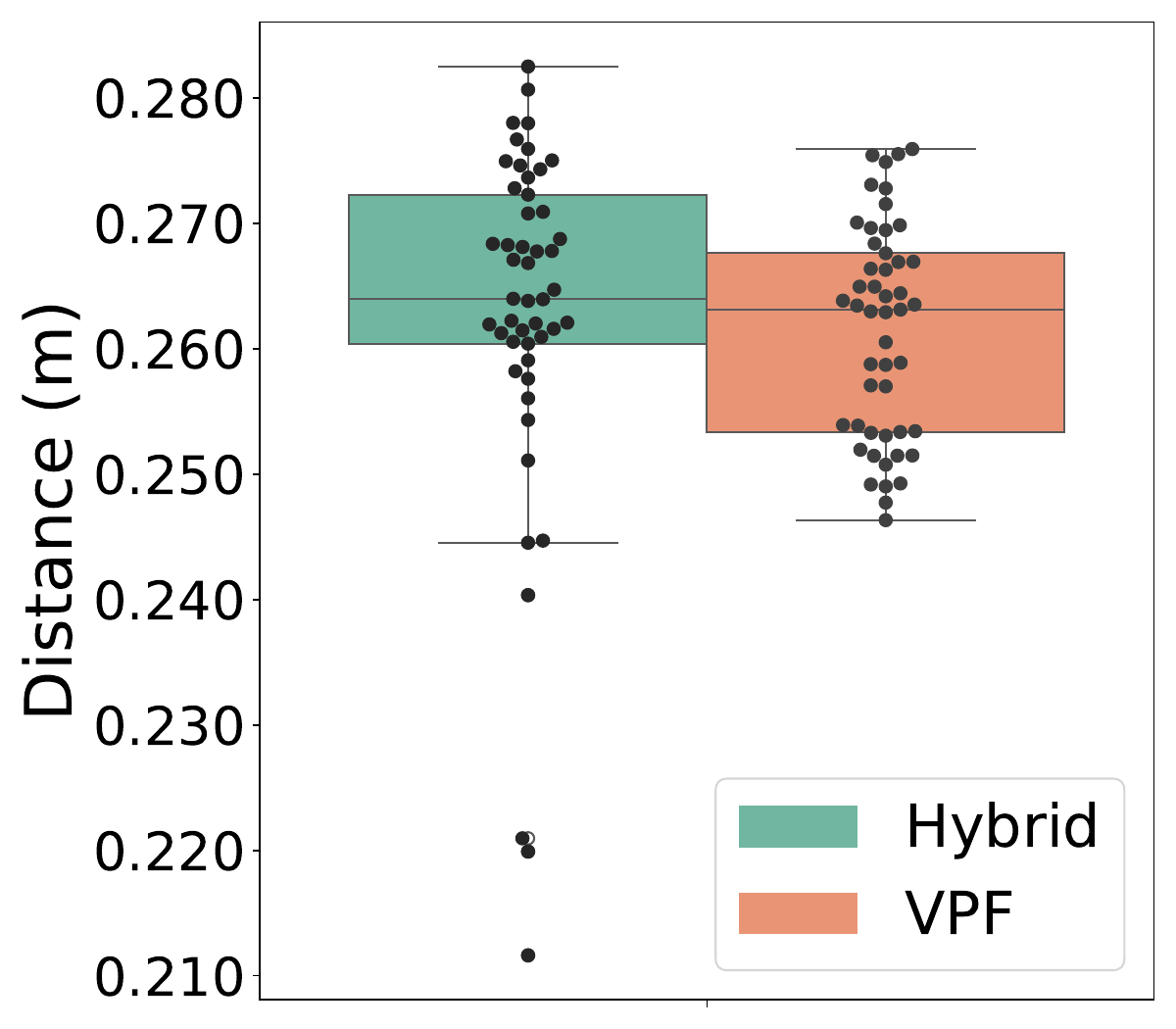}} & 
        \hspace{1mm} % Adds a number of mm of horizontal space
        \scalebox{1.0}[1.1]{\includegraphics[width=0.19\textwidth]{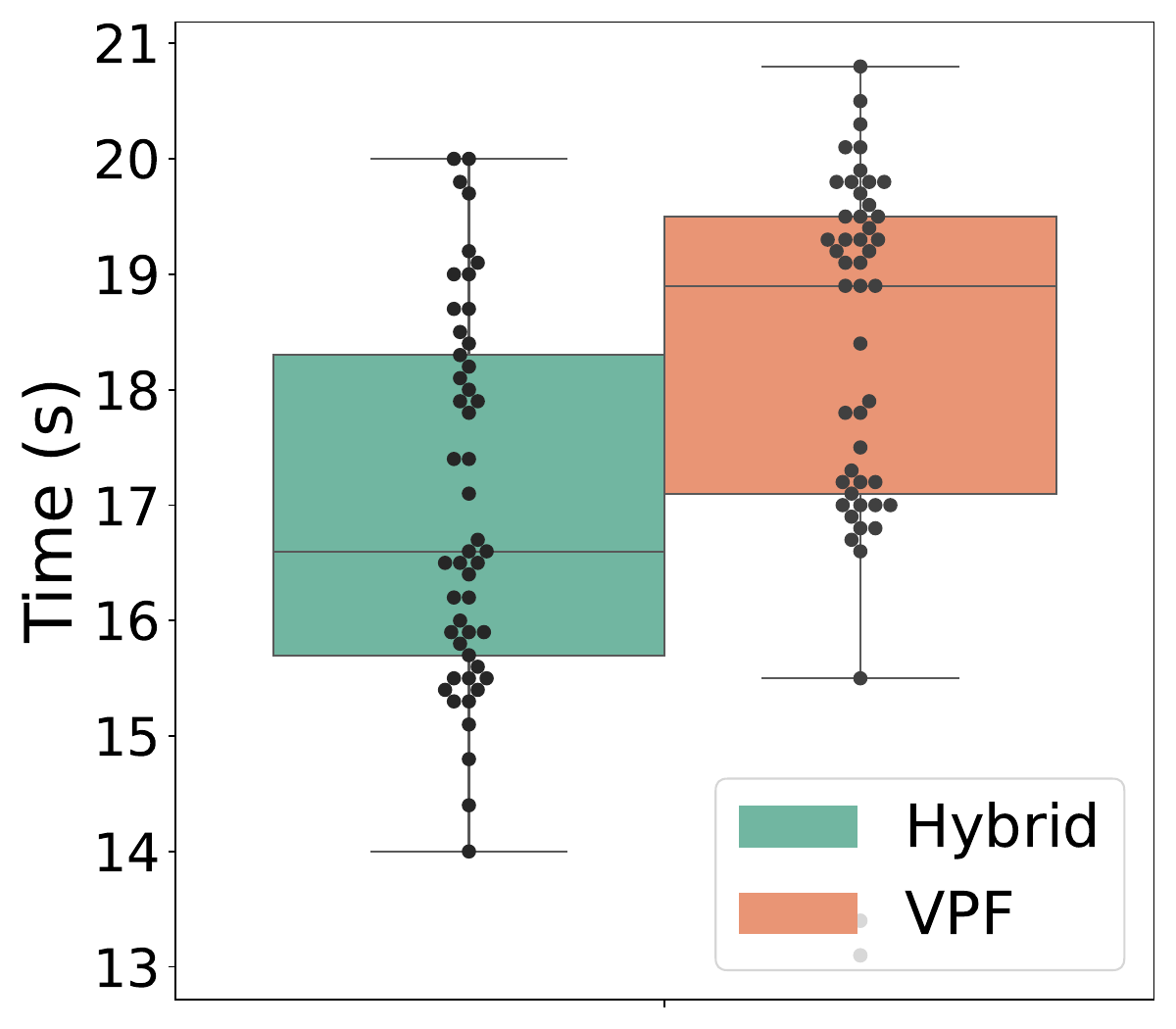}} & 
        \hspace{1mm}
        \scalebox{1.0}[1.1]{\includegraphics[width=0.19\textwidth]{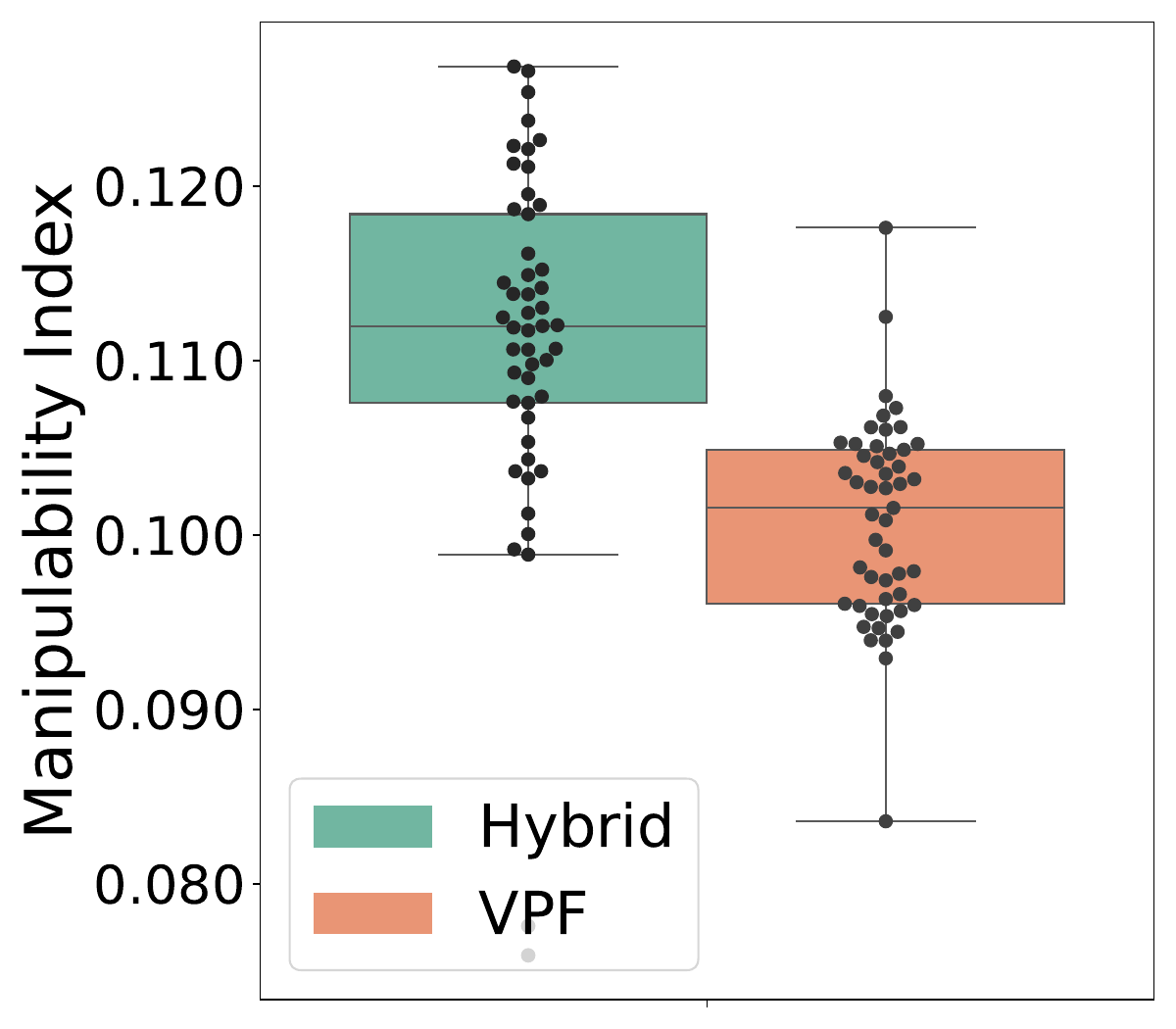}} & 
        \hspace{1mm}
        \scalebox{1.0}[1.1]{\includegraphics[width=0.19\textwidth]{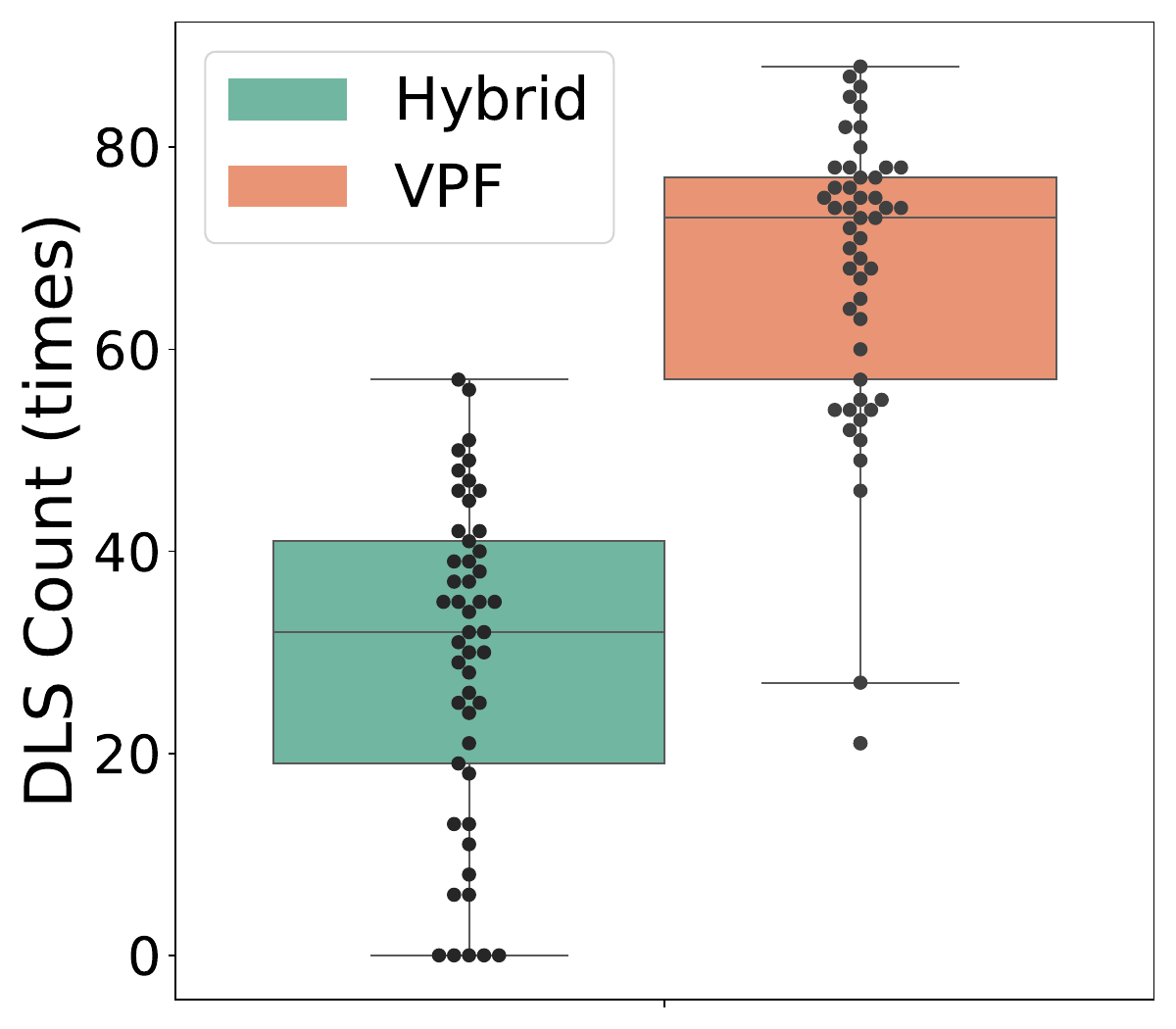}} & 
        \hspace{1mm}
        \scalebox{1.0}[1.1]{\includegraphics[width=0.19\textwidth]{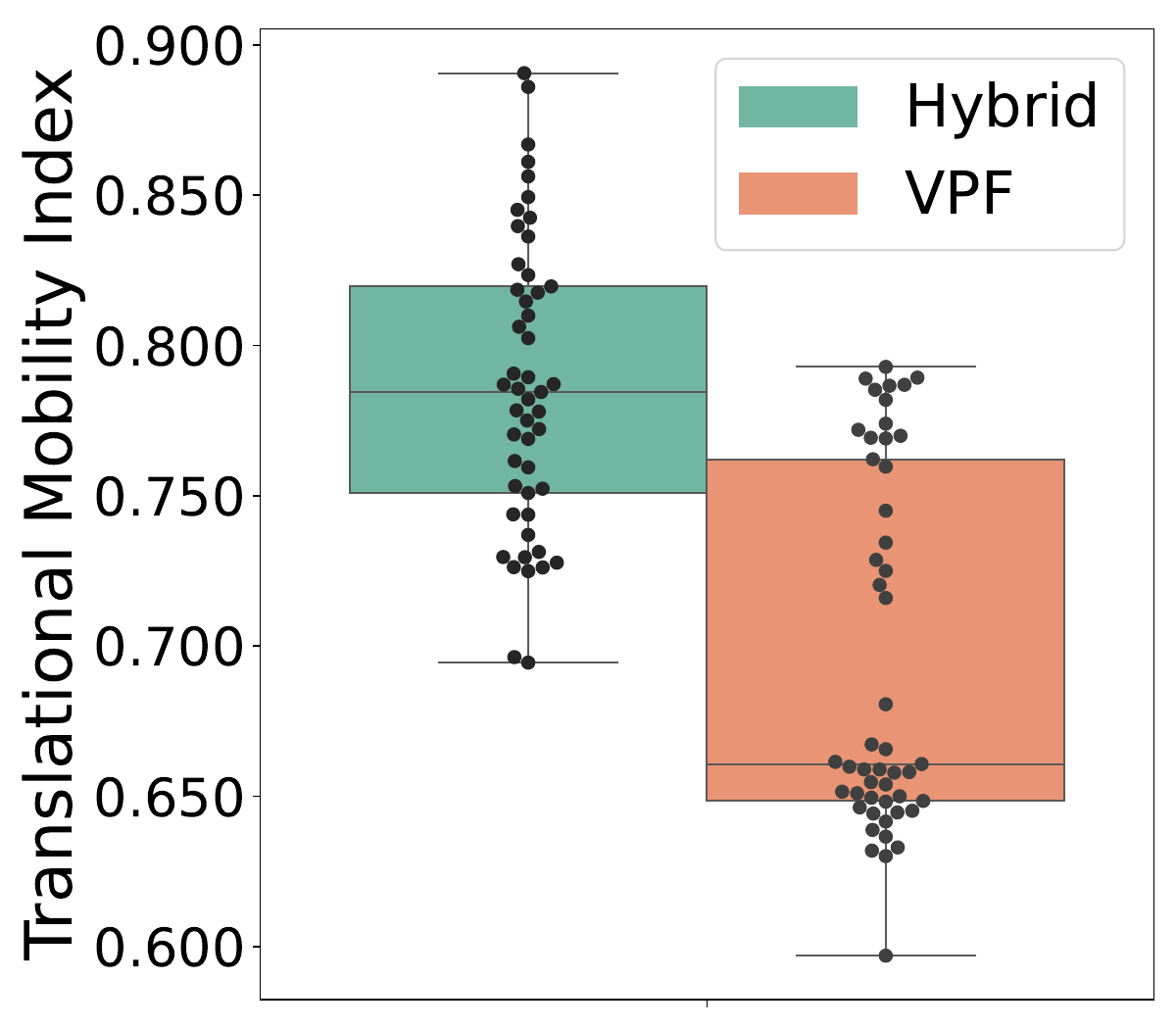}} \\
        \footnotesize (a) Closest distance & 
        \hspace{1mm}
        \footnotesize (b) Time & 
        \hspace{1mm}
        \footnotesize (c) Average manipulability & 
        \hspace{1mm}
        \footnotesize (d) DLS count & 
        \hspace{1mm}
        \footnotesize (e) Translation mobility
    \end{tabular}
    \caption{Hybrid and VPF comparison after 50 runs, showing metrics for minimum distance to obstacles, total time taken, average manipulability, count of DLS adjustments, and average translation mobility ratio, which provide a comprehensive evaluation of performance and efficiency.}
    
    \label{fig:multiple_figures}
\end{figure*}

Fig.~\ref{fig:multiple_figures} compares the hybrid and VPF planners across 50 runs, giving both successfully navigated to the destination without collisions. The metrics include minimum distance to obstacles, time to destination, average manipulability, DLS intervention frequency, and translation mobility ratio. A paired t-test for each category outputs a corresponding $p$-value, which is compared to a significance level of 0.05 to determine the statistical significance of the differences. 
Both methods give statistically analogous results for average distance to obstacles ($\approx 0.262, p = 0.700 $) demonstrating equivalent safety through balanced attractive and repulsive effects. However, the hybrid planner guides the robot to the goal faster than the VPF planner(17.001s vs 18.322s, $p\ll\lambda$), benefiting from its more efficient global path, especially when obstacles do not heavily obstruct the path.
The hybrid planner achieves higher average manipulability (0.112 vs 0.101,  $p\ll\lambda$), resulting in fewer DLS applications (30 vs 68 times, $p\ll\lambda$). It also achieves a higher average translational mobility ratio (0.787 vs 0.696, $p\ll\lambda$), indicating improved efficiency in the task-space moving direction.
These results showcase the hybrid planner's ability to combine global path optimisation with reactive local planning, leading to improved safety, efficiency, and manipulability in dynamic environments.

\subsection{Real hardware}

In this section, a set of experiments is conducted on the Sawyer robot implemented the hybrid planner, to move from start to goal configurations, under random movement of a box obstacle. The box is tracked using the Motion Capture (MoCap) Optitrack system. The primary objective of this experiment is to verify the hybrid planner in a real-world scenario, where the manipulator can avoid the moving obstacle and complete the task with feasible movements. 

The first experiment is with a static obstacle placed in the environment after the pre-planning phase, indicating the global path does not contain information about the obstacle, and the avoidance movement totally relies on the local planning phase. Fig.~\ref{fig:ex_real_annotated_0} demonstrates a run in this experiment, where the red line shows the end effector's position in the pre-planned global path, and the green line is its actual position during the robot's movement. Since the box is kept static, the Sawyer silhouettes can overlap in the figure to illustrate its adaptive motion to avoid the box. From this figure, it can be seen that although the task is to track an obstructed path, the hybrid planner can still help the robot avoid unexpected object during the movement and also guide it successfully to the goal position.
%
% \begin{figure}%[htbp]
% \centerline{\includegraphics[width=0.95\linewidth]{experiment_real_with_annotations_0_colour_fixed.pdf}}
% \caption{Trajectory executed on a Sawyer robot with a static obstacle}
% \label{fig:ex_real_annotated_0}
% \end{figure}

\begin{figure}%[htbp]
\centerline{\includegraphics[width=0.95\linewidth]{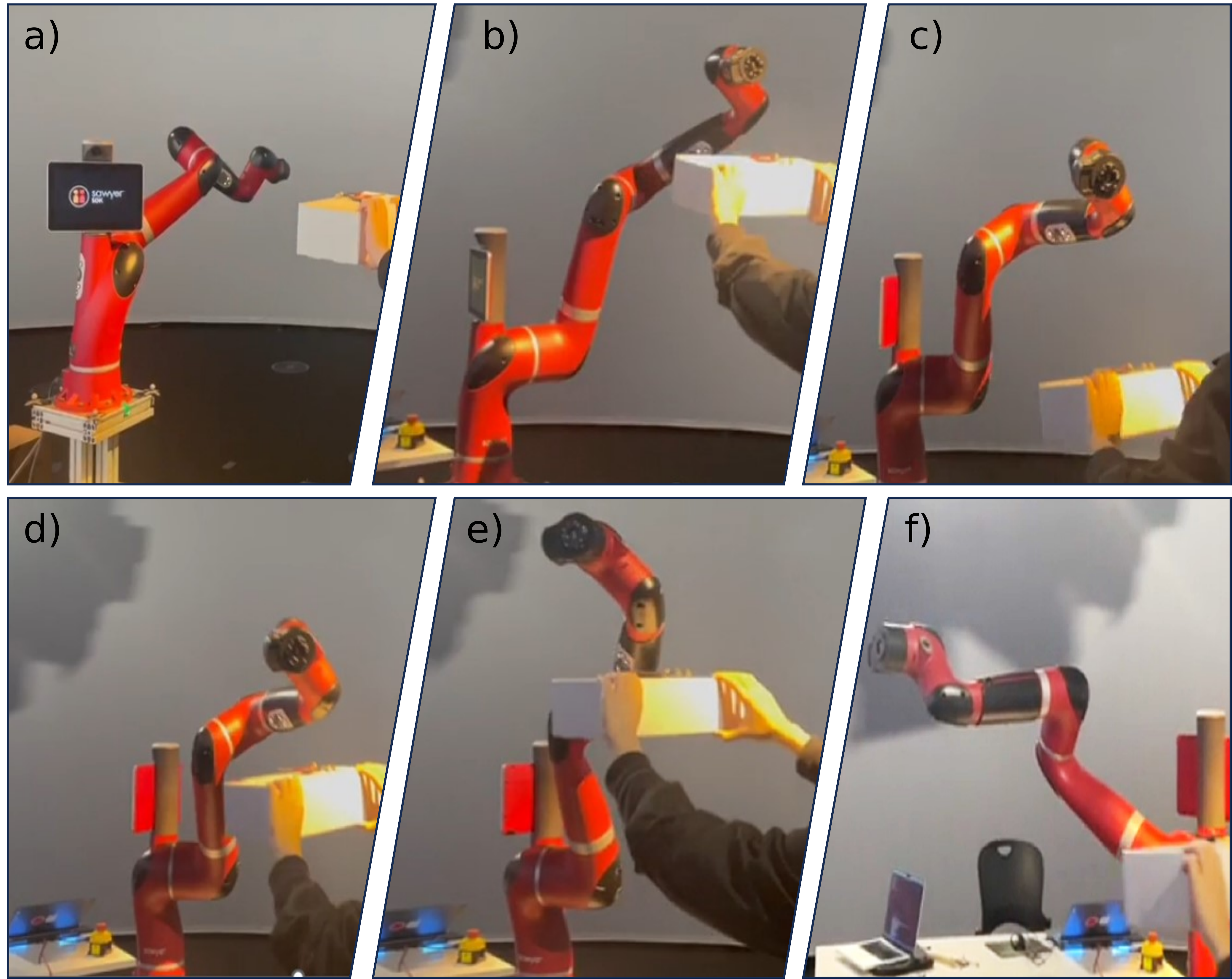}}
\caption{Configurations of the Sawyer robot and box positions during a task. Images (a)-(f) show the box's movement sequence and the robot's reactionary movements for a single run.}
\label{pe11}
\end{figure}

\begin{figure}%[htbp]
% \centerline{\includegraphics[width=1\linewidth]{pe1.pdf}}
\centerline{\includegraphics[width=0.85\linewidth]{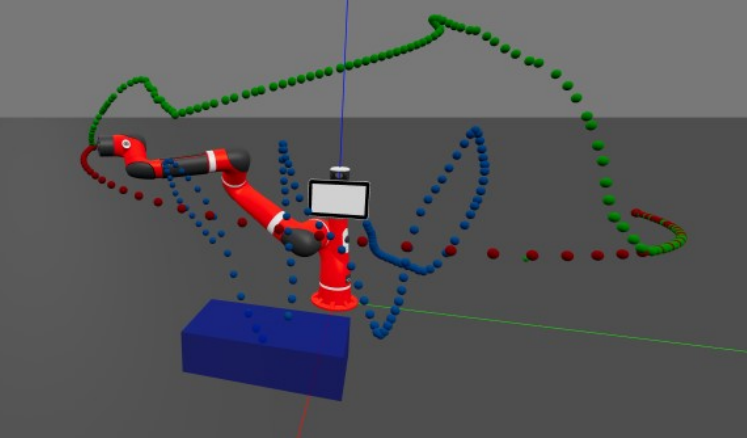}}
\vspace{2mm} % Adds a few mm of vertical space
\centerline{\includegraphics[width=0.95\linewidth]{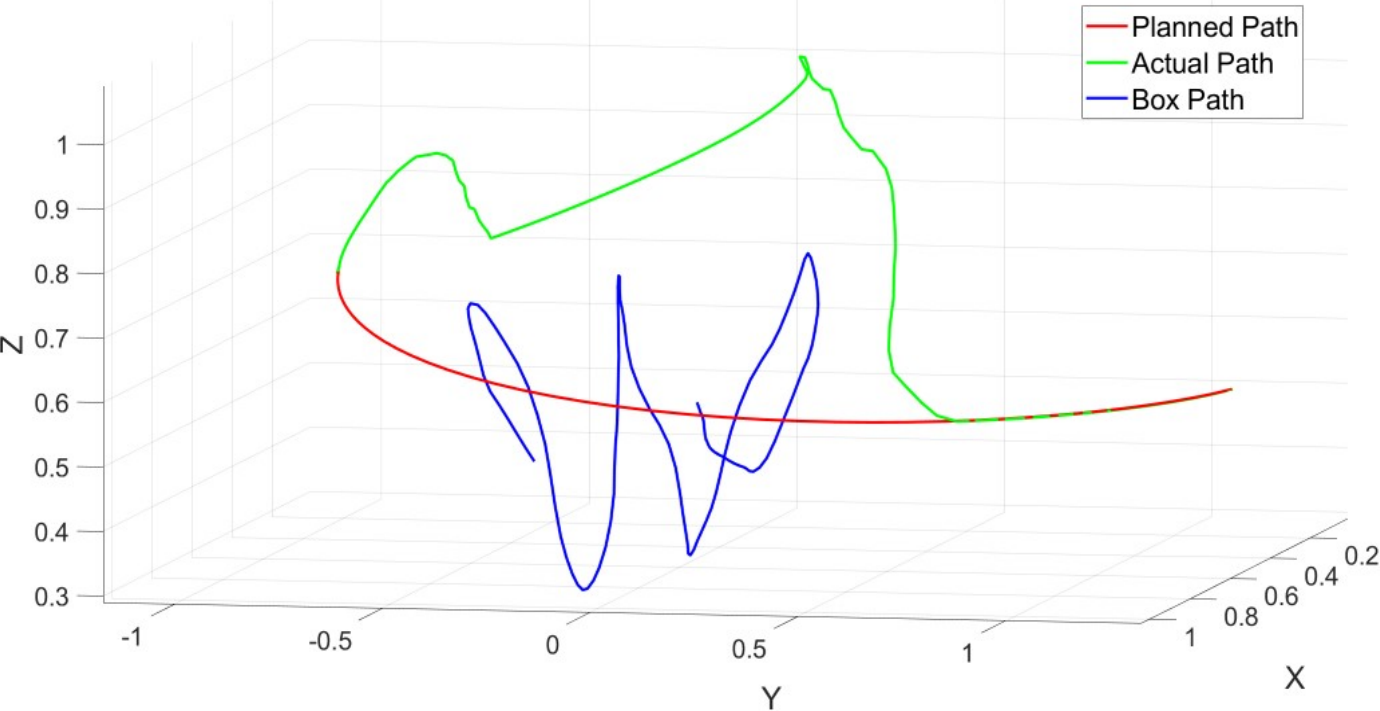}}
\caption{Visualisation of the collected motion data in Fig.~\ref{pe11}. Top: simulation environment. Bottom: plotted paths.}
\label{pe1}
\end{figure}

\begin{figure}[t]
    \centering
    \begin{tabular}{c}
         \scalebox{1.0}[1.1]{\includegraphics[width=0.42\textwidth]{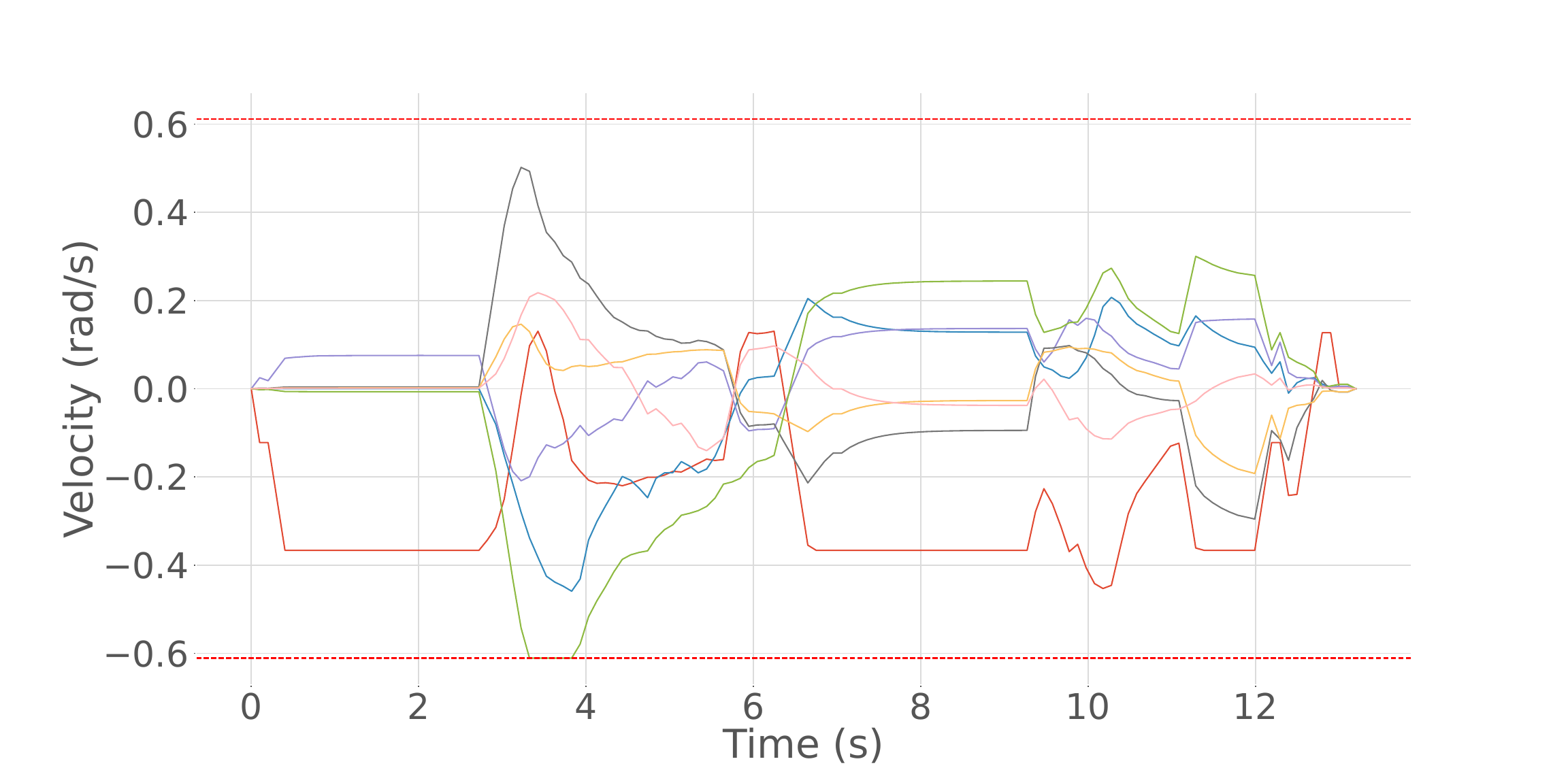}}\\
        \scalebox{1.0}[1.1]{\includegraphics[width=0.42\textwidth]{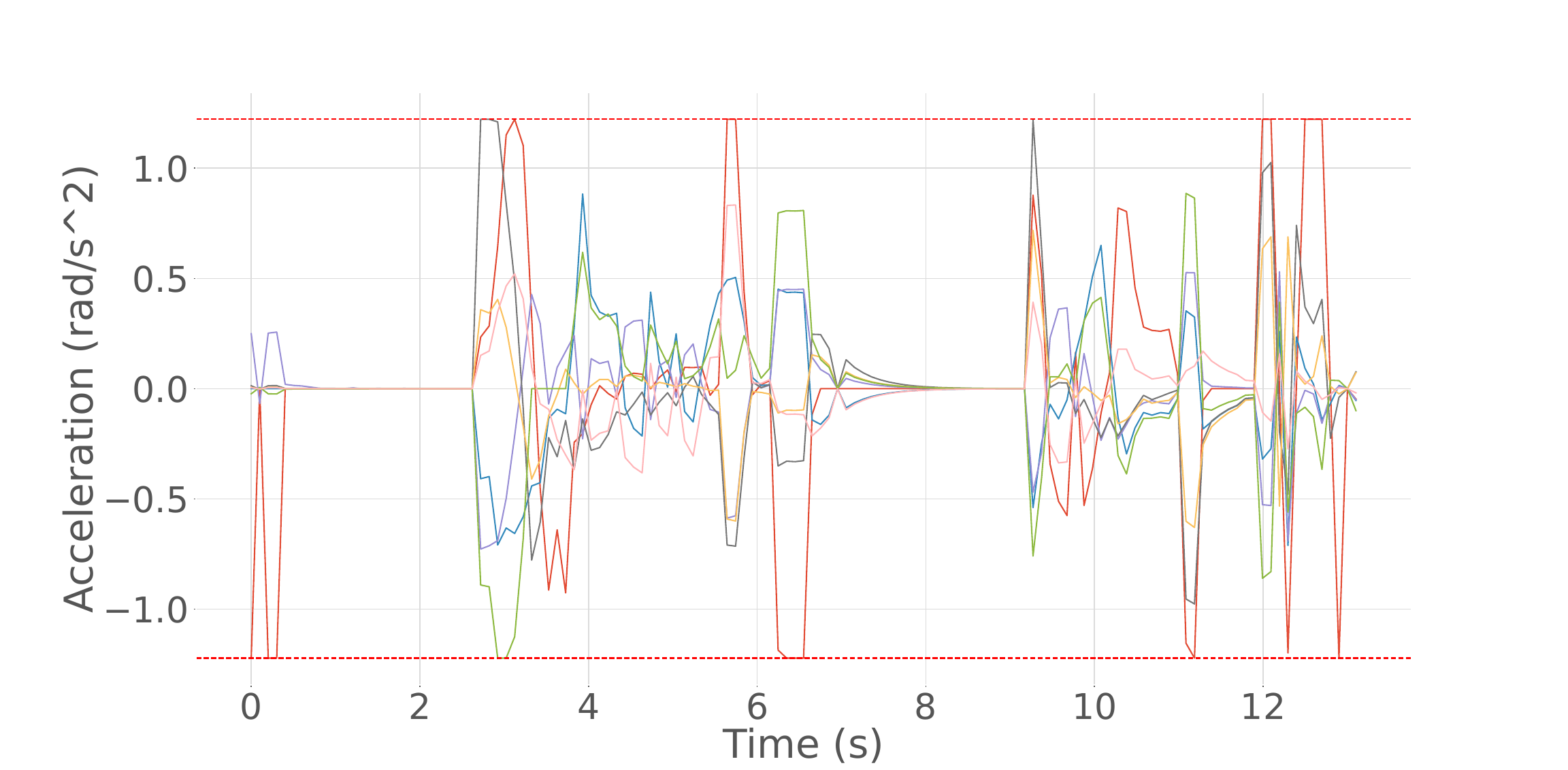}}\\
    \end{tabular}
    \caption{Motion profiles for each Sawyer's joint from the collected motion data in Fig.~\ref{pe11}. Top: joint velocity. Bottom: joint acceleration. The red dashed reference lines are for the velocity and acceleration limits.}
    \label{pe2}
\end{figure}

The second experiment also has an obstacle-free global path, and the box is introduced to the environment after the robot starts to move. However, this time, the box is randomly moved within the Sawyer workspace. Fig.~\ref{pe11} is a run time in this experiment, that shows the Sawyer reactions in the relationship with different box positions during their movements. Data collected from Fig.~\ref{pe11} has their corresponding visualisation in Fig.~\ref{pe1}, with different coloured lines for the box positions, end effector's preplanned and actual motions. Fig.~\ref{pe2} is the velocity and acceleration profile for each robot joint, showing that all the joint motions are bounded within constraints.

%
% \begin{figure}%[htbp]
% \centerline{\includegraphics[width=1\linewidth]{./Figures/pe_vel.pdf}}
% % \vspace{3mm} % Adds a few mm of vertical space
% \centerline{\includegraphics[width=1\linewidth]{./Figures/pe_acc.pdf}}
% \caption{Motion profiles for each Sawyer's joint from the collected motion data in Fig.~\ref{pe11}. Top: joint velocity. Bottom: joint acceleration. The red dashed reference lines are for the velocity and acceleration limits.}
% \label{pe2}
% \end{figure}

% \begin{figure}%[htbp]
% \centerline{\includegraphics[width=1\linewidth]{p12.pdf}}
% \caption{Experiment 2}
% \label{p12}
% \end{figure}

% \begin{figure}%[htbp]
% \centerline{\includegraphics[width=1\linewidth]{pe2.pdf}}
% \caption{Experiment 2 Visualisation}
% \label{pe2}
% \end{figure}

% -------------------------------------------------

\section{Evaluation and Discussion}

The experimental results demonstrate that the hybrid planner outperforms traditional VPF in environments with moderately dynamic obstacles, particularly in tasks requiring predefined criteria, such as enhancing manipulability or shorter execution times. This advantage stems from the combination of global path optimization and local reactive planning. However, the effectiveness of the hybrid planner is highly dependent on the quality of the initial global path and the degree of environmental change. In highly dynamic environments where obstacles significantly obstruct the pre-planned path, the planner may struggle to maintain optimal performance. The hybrid planner, therefore, is best suited for scenarios with predictable or semi-structured dynamic environments. It shows promise in applications such as manufacturing for assembly tasks with anticipated component movements, warehousing for item sorting and stacking in changing but relatively stable layouts, and minimally invasive surgery for adapting to patient-specific anatomies and minor organ shifts. In these contexts, the SBMP component provides an efficient global strategy in a short time, while the VPF component offers real-time adaptability to local changes. This combination enables efficient and safe motion planning in complex, dynamic environments where complete environmental information is not always available.

\section{Conclusion}

This study presents an efficient, cost-effective hybrid planner that enhances motion planning for manipulators in dynamic and uncertain environments. By combining an SBMP for global path generation with a VPF method for local obstacle avoidance, the hybrid approach leverages the strengths of both techniques. This integration results in a more efficient and adaptable planning strategy that reduces the need for frequent replanning and extensive environmental information. The hybrid planner demonstrates improved performance in navigating cluttered environments, particularly in scenarios with multiple obstacles and where an initial optimal path can be established. Experimental results show enhanced manipulability, faster execution times, and better obstacle avoidance compared to traditional VPF methods. These improvements make the hybrid planner particularly suitable for applications in semi-structured dynamic environments, such as advanced manufacturing, warehouse automation, and minimally invasive surgical procedures.
Future research directions should focus on further improving the planner's adaptability to highly unpredictable environments. This could involve developing methods for dynamic global path updates, integrating online learning techniques, and exploring new combinations of global and local planning strategies. As robotic applications continue to expand into more complex and dynamic domains, the principles demonstrated in this hybrid approach offer a promising foundation to address the evolving challenges of robot motion planning.

% In conclusion, this study introduces a hybrid planner as an innovative, cost-effective solution for enhancing motion planning in robotic manipulators navigating dynamic and uncertain environments. The hybrid planner strategically combines the strengths of an SBMP as a global planner with the local planning capabilities of the VPF for obstacle avoidance. By relying on an accurately preplanned, asymptotically optimal path generated by the SBMP, the hybrid approach mitigates the complex and resource-intensive replanning processes typically associated with SBMP and APF. Simultaneously, it minimises the requirement for exhaustive environmental information and navigates cluttered environments more efficiently than VPF, particularly in scenarios with numerous obstacles and where an optimal global path for robot operation is already established. Therefore, the hybrid planner is well-suited for applications in dynamic and uncertain environments where object positions are subject to change, such as warehouse autonomous or medical robotics for minimally invasive surgery.

% Looking towards future research directions, there are opportunities to explore diverse combinations of global and local planners, optimising motion planning strategies for manipulators. The insights from this study underline the hybrid planner's potential as an adaptable solution for addressing the challenges of motion planning in dynamic environments, thereby contributing to the ongoing evolution of robotics in diverse application domains.

\section*{Acknowledgements}

The authors acknowledge the UTS:RI for providing the essential resources and laboratory facilities to conduct this research. This work was supported by the ARC Industrial Transformation Training Centre (ITTC) for Collaborative Robotics in Advanced Manufacturing under grant IC200100001. D.T. Le and D.D.K. Nguyen are recipients of Australian Government Research Training Program Scholarships. Special thanks are extended to Chris Hancock, Anh Minh Tu, and Mai Thao Trinh for their valuable technical assistance and unwavering support throughout the study.

\vspace{-30pt}
\begin{IEEEbiography}
[{\includegraphics[width=1in,height=4in,keepaspectratio]{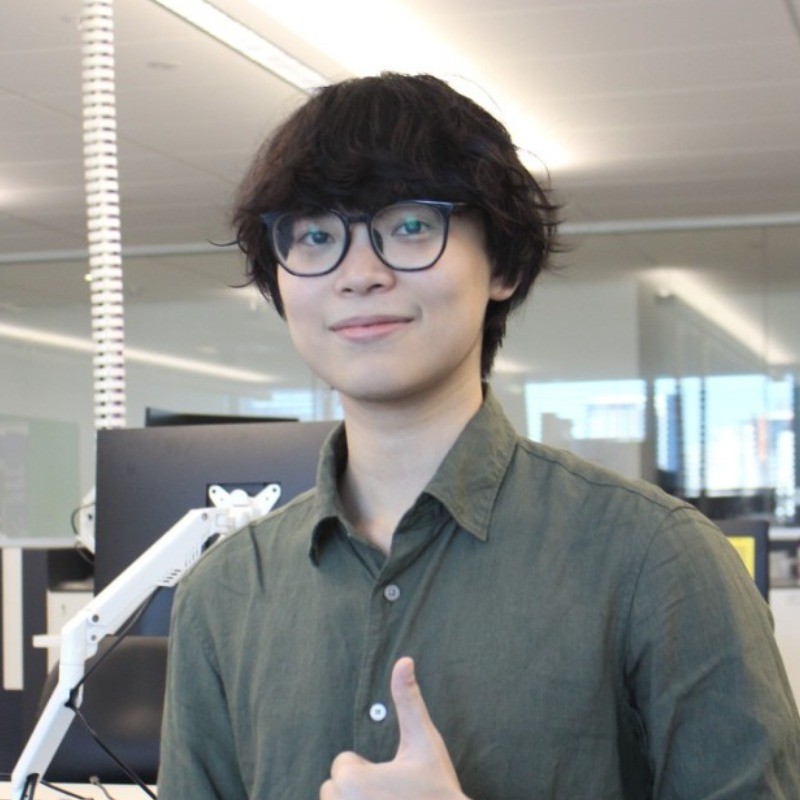}}]{Ho Minh Quang Ngo} received the Bachelor of Engineering degree in Mechatronics Engineering from the University of Technology Sydney, Sydney, Australia. He is currently working at the Robotics Institute, University of Technology Sydney, Sydney, Australia. His research interests include control, planning and reinforcement learning for robotics system.
\end{IEEEbiography}
\vspace{-40pt}
\begin{IEEEbiography}
[{\includegraphics[width=1in,height=3in,clip,trim=0 0 0 1.5in,keepaspectratio]{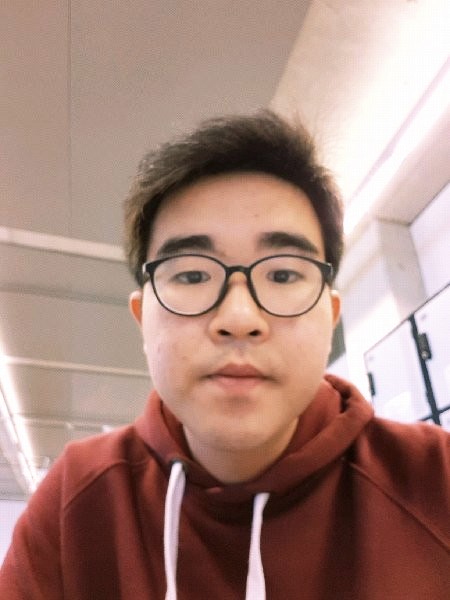}}]{Dac Dang Khoa Nguyen} (Student Member, IEEE) received the Bachelor of Engineering degree in Mechatronics Engineering from the University of Technology Sydney, Sydney, Australia. He is currently a PhD candidate with the Robotics Institute, University of Technology Sydney, Sydney, Australia. His research interests include multi-robot system control, planning, and decision making.
\end{IEEEbiography}
\vspace{-40pt}
\begin{IEEEbiography}
[{\includegraphics[width=1in,height=4in,clip,keepaspectratio]{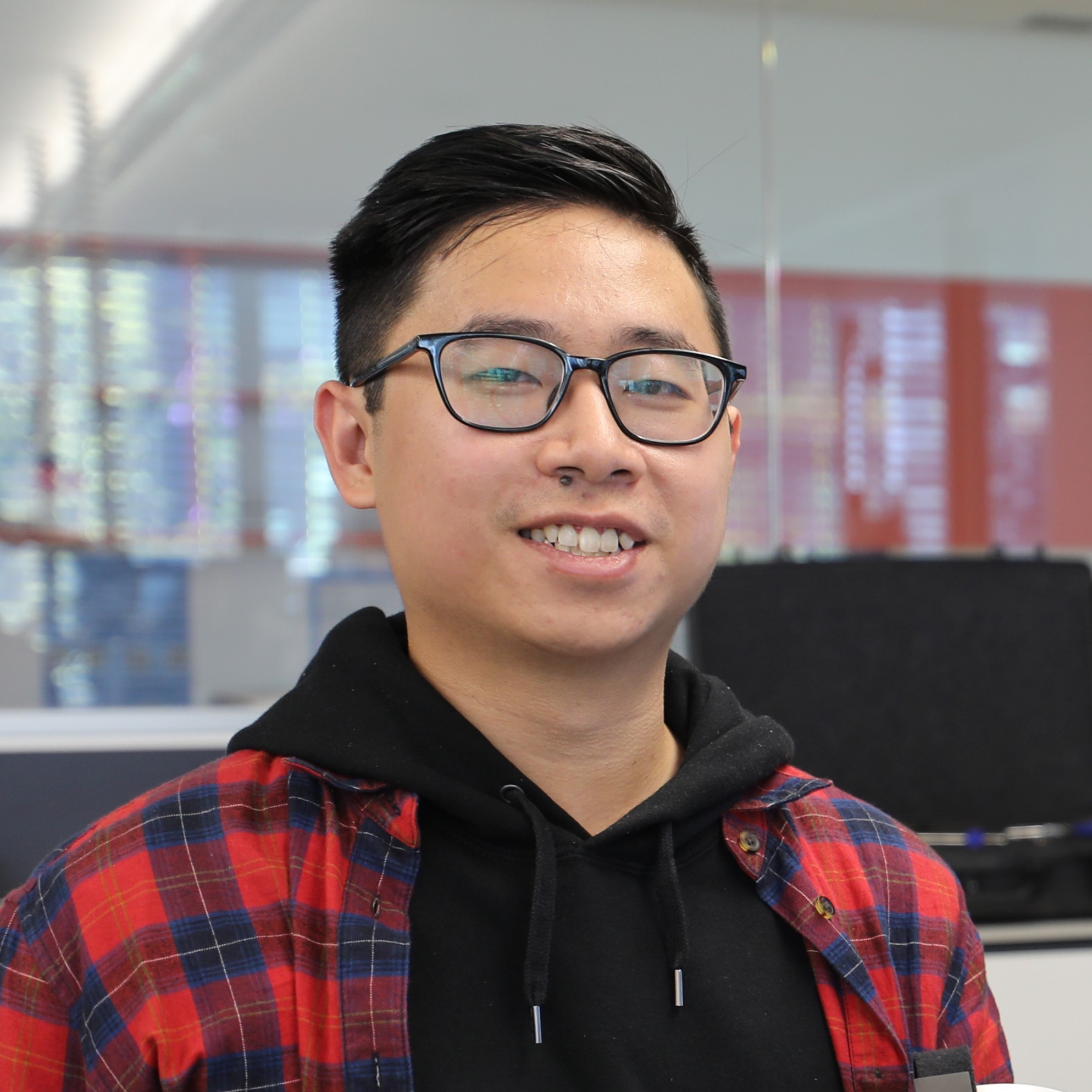}}]{Dinh Tung Le} (Student Member, IEEE) received the Bachelor of Engineering degree in Mechatronics Engineering from the University of Technology Sydney, Sydney, Australia. He is currently a PhD candidate with the Robotics Institute, University of Technology Sydney, Sydney, Australia. His research interests include XR technologies for Human-Robot Interaction.
\end{IEEEbiography}
\vspace{-40pt}
\begin{IEEEbiography}
[{\includegraphics[width=1in,height=4in,clip,keepaspectratio]{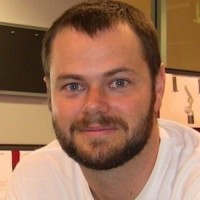}}]{Gavin Paul} (Member, IEEE) received the Bachelor of Engineering, Bachelor of Arts, Computer Systems Engineering from the University of Technology Sydney, Sydney, Australia and the Ph.D degree in Mechatronics, Robotics and Automation Engineering from the University of Technology Sydney, Sydney, Australia. He is currently an Associate Professor at the University of Technology Sydney, Sydney, Australia. His research interests include artificial intelligence, control engineering, mechatronics and robotics, and manufacturing engineering.
\end{IEEEbiography}

\end{document}